# Graph Neural Networks for Quantifying Compatibility Mechanisms in Traditional Chinese Medicine

Jingqi Zeng[1], Xiaobin Jia[1,2*]

[1]School of Traditional Chinese Pharmacy, China Pharmaceutical University; Nanjing, China.

[2]State Key Laboratory of Natural Medicines, China Pharmaceutical University; Nanjing, China.

*Corresponding author. Email: jiaxiaobin2015@163.com

**Abstract:** Traditional Chinese Medicine (TCM) involves complex compatibility mechanisms characterized by multi-component and multi-target interactions, which are challenging to quantify. To address this challenge, we applied graph artificial intelligence to develop a TCM multi-dimensional knowledge graph that bridges traditional TCM theory and modern biomedical science (https://zenodo.org/records/13763953 ). Using feature engineering and embedding, we processed key TCM terminology and Chinese herbal pieces (CHP), introducing medicinal properties as virtual nodes and employing graph neural networks with attention mechanisms to model and analyze 6,080 Chinese herbal formulas (CHF). Our method quantitatively assessed the roles of CHP within CHF and was validated using 215 CHF designed for COVID-19 management. With interpretable models, open-source data, and code (https://github.com/ZENGJingqi/GraphAI-for-TCM ), this study provides robust tools for advancing TCM theory and drug discovery.

## One-Sentence Summary

We quantified Traditional Chinese Medicine compatibility using graph artificial intelligence, enabling interpretable insights into herbal synergy.

# Graphic Abstract

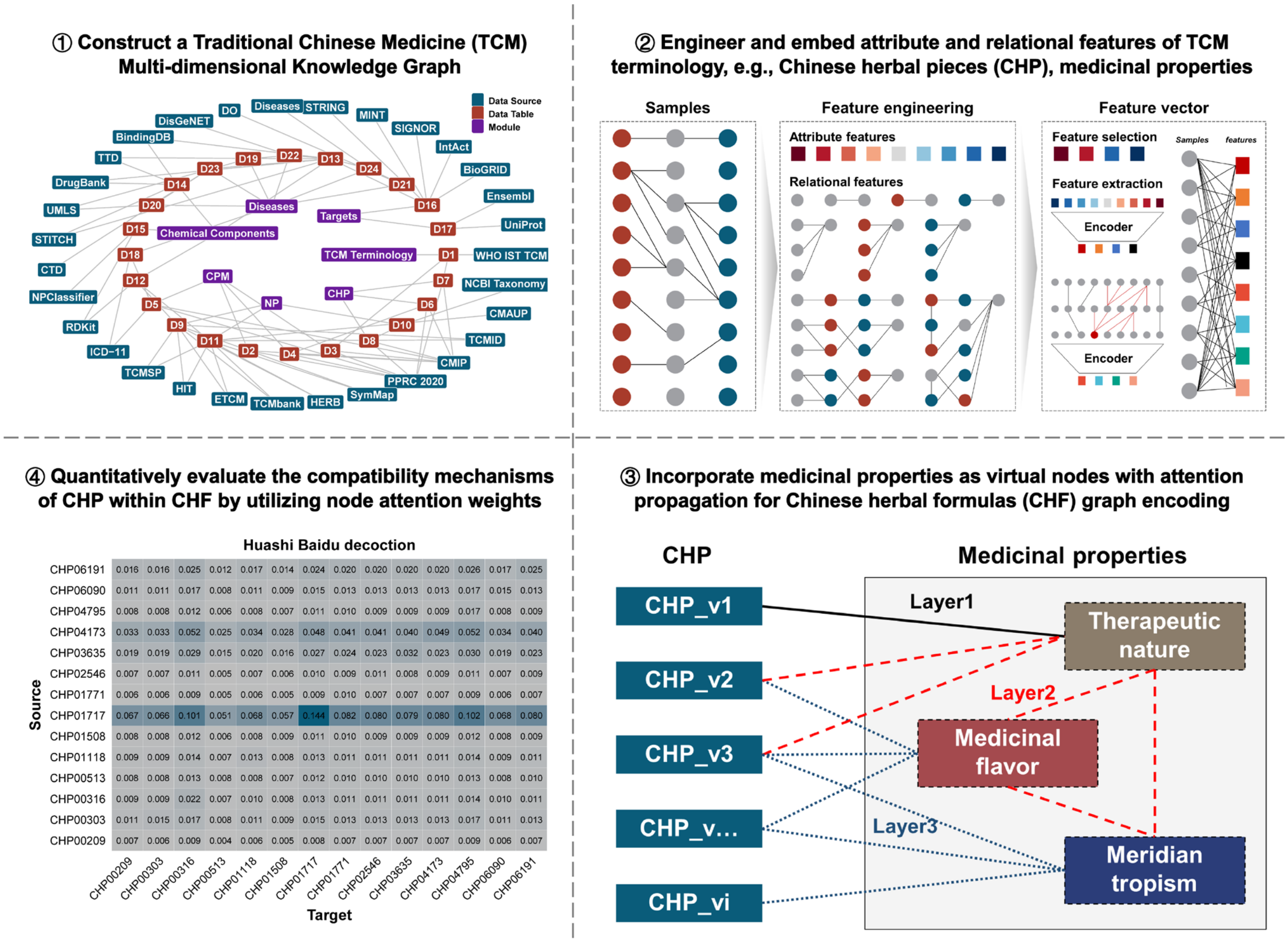

# Highlights

1) Constructed a multi-dimensional TCM knowledge graph that integrates traditional concepts with modern biomedical data.

2) Introduced virtual nodes in graph encoding to improve the modeling of compatibility mechanisms in Chinese herbal formulas.

3) Provided open-source data, models, and code to support TCM research and practical applications.

4) Quantified the roles of key Chinese herbal pieces and herb pairs in the management of COVID-19.

## Introduction

Traditional Chinese Medicine (TCM) emphasizes a holistic perspective and dynamic balance, focusing on the comprehensive evaluation of an individual's constitution, disease state, and the interplay among various physiological systems. Through the rational compatibility of herbal medicines, TCM achieves synergistic effects and toxicity reduction (*1*). This core principle has demonstrated significant advantages in treating complex diseases. For example, Lianhua Qingwen Capsules have shown remarkable efficacy in alleviating symptoms and reducing hospitalization time for COVID-19 patients (*2*, *3*). Additionally, PHY906, developed by Yale University based on the traditional Huangqin Decoction, has improved colorectal cancer treatment outcomes as a chemotherapy adjuvant (*4*).

In recent years, the rapid development of artificial intelligence (AI) has introduced novel opportunities for investigating the complex mechanisms underlying TCM (*5*, *6*). AI's exceptional data processing capabilities, particularly in multi-dimensional data analysis and complex relationship modeling, are transforming traditional medicine from experience-driven to data-driven paradigms (*7–9*). Notably, Graph Artificial Intelligence (GraphAI) offers a unique toolkit for exploring complex network-structured data by integrating knowledge graphs, graph computation, and graph neural networks (GNNs) (*10*, *11*). The core challenges of TCM compatibility—complex interactions involving multiple components, targets, and pathways—align closely with GraphAI's strengths in handling intricate relationships (*12–14*).

However, existing data-driven approaches in TCM research face three major challenges. First, the quantification and analysis of intricate herbal interactions remain insufficient, with existing models lacking interpretability to provide clear mechanistic insights. Second, the design of current models often fails to incorporate core TCM theories, limiting the alignment between theoretical foundations and computational frameworks. Lastly, the absence of high-quality, multi-dimensional, and openly shared TCM datasets constrains interdisciplinary collaboration and innovation.

To address these challenges, this study proposes a method for quantifying TCM compatibility mechanisms based on GraphAI, deeply integrating TCM theories with modern AI technologies to achieve the digitization and quantification of TCM compatibility. We developed a TCM Multi-dimensional Knowledge Graph (TCM-MKG), encompassing multi-layered information spanning traditional TCM theories and modern biomedical science. This graph integrates TCM compatibility relationships, chemical compositions, target interactions, biological networks, and disease-diagnosis knowledge, offering comprehensive multi-dimensional data support for elucidating TCM's complex systems.

Using the TCM-MKG, we modeled Chinese herbal pieces (CHP) as units, incorporating medicinal properties and diagnostic theories. By leveraging graph neural networks with attention mechanisms, we analyzed 6,080 Chinese herbal formulas (CHF) and constructed highly interpretable models to quantify the compatibility roles of CHP within CHF. This method unveiled synergistic effects and compatibility mechanisms among herbs and was validated on 215 CHF used in China's COVID-19 management from 2019 to 2021. The results demonstrate the method's efficacy in advancing TCM theory modernization, data-driven research, and drug discovery, providing new pathways for disciplinary progress and enhanced healthcare services.

# Results

### *Data structure and compound-target discovery in the TCM multi-dimensional knowledge graph*

Data-driven TCM research has long been hindered by challenges in data standardization and interdisciplinary integration, limiting the synthesis of traditional theories with modern biomedical knowledge. To address these issues, we developed the TCM Multi-Dimensional Knowledge Graph (TCM-MKG), integrating seven core modules: TCM terminology, Chinese patent medicines (CPM), Chinese herbal pieces (CHP), natural products (NP), chemical compounds, targets, and diseases (**Figure 1A**). Data from 31 authoritative databases were standardized via semantic integration, forming a multi-layered framework comprising 24 primary data tables (detailed in **Data S1**), enabling systematic interdisciplinary research.

The TCM terminology module includes 1,810 standardized terms foundational for understanding etiology, pathogenesis, and therapeutic principles. These terms are linked to CPM, disease, and target modules, forming the semantic core of the knowledge graph. The CPM module documents 8,977 formulas, connecting them to TCM terms (11,185 records) and CHPs (74,084 records). Additionally, 69,431 links to ICD-11 codes ensure interoperability with Western medicine. On average, CPM formulas comprise eight CHPs (**Figure 1B**), with a mean dose ratio of 0.11 among CHPs (**Figure 1C**), reflecting the typical multi-herbal compatibility of TCM.

The CHP module describes 6,207 CHPs with 23,517 medicinal property records and 226,589 associations with chemical compounds, elucidating the pharmacological basis of CHPs. The NP module includes 4,810 natural products, linked to chemical compounds through 426,305 associations. Despite these advances, only 12.1% of the 123,647 documented compounds had defined targets (**Figure 1D**), creating a bottleneck in studying molecular mechanisms.

To overcome this limitation, we developed a neighbor-diffusion-based compound-target prediction strategy using graph neural networks (GNNs). Molecular structures were represented as graphs using RDKit, with atoms and bonds encoded as nodes and edges, respectively. We employed three embedding methods—Graph Convolutional Network (GCN), Graph Attention Network (GAT), and Graph Autoencoder (GAE)—combined with 197 molecular descriptors and 2,048-bit Morgan fingerprints to capture physicochemical and structural properties. Uniform Manifold Approximation and Projection (UMAP) reduced high-dimensional features, preserving local and global data structures (**Figure 1E**). Candidate compound-target pairs were generated using a neighbor-diffusion approach, producing 7,260,578 extended target associations (**Figure 1F**), validated by the PSICHIC model (*15*). Filtering for high-confidence predictions using a binding affinity threshold of 5.0 ensured reliability.

Binding affinity analysis revealed that "TCM diffused targets" matched the high-affinity range (>5.0) distribution of "High fidelity targets," while "Recorded TCM targets" exhibited lower affinities. This confirmed the effectiveness of the neighbor-diffusion strategy. Threshold analyses showed significant coverage improvements (**Figures 1G–I**): InChIKey: Coverage rose to 99.86%, with compounds increasing from 7,631 to 62,744. EntrezID: Coverage reached 99.0%, with targets increasing from 13,781 to 14,370. Compound-target associations: Expanded from 343,373 to 6,526,940, an 18-fold increase.

Further analysis of feature representation methods, molecular weight, and target sequence length confirmed the reliability and broad applicability of this approach (**Figures S1, S2**, Supporting

Information). This work significantly enhances the compound-target association landscape in TCM, providing a robust foundation for investigating compatibility mechanisms. All extended datasets are publicly available at https://zenodo.org/records/13763953 .

### *Feature engineering and embedding of Chinese herbal pieces*

The multi-dimensional characteristics of CHPs introduce complexity into pharmacological research. To comprehensively represent their origin, medicinal properties, efficacy, and compatibility, we proposed a multi-modal feature integration method. By combining feature engineering with vector embedding techniques, unified feature vector representations were generated. Inspired by multi-modal feature integration and multi-task learning strategies (*16*), this approach provides a robust foundation for analyzing CHPs' roles in complex compatibility mechanisms.

**Figure 2** outlines the encoding process for multi-source CHP features and their dimensionality-reduction visualizations. For origin features, we extracted 10 principal components using phylogenetic tree genetic distances and principal component analysis (PCA). These were combined with dummy variables for taxonomic kingdoms (Plantae, Animalia, Fungi, Algae, and Mineralia) to create a 15-dimensional origin feature vector. The extraction and phylogenetic structure of these features are detailed in **Figure S3** (Supporting Information). Relational attributes—medicinal property, efficacy, and compatibility features—were encoded using a Word2Vec model, producing 15-dimensional medicinal property vectors, 30-dimensional efficacy vectors, and 30-dimensional compatibility vectors. To ensure consistency, all features were standardized before integration (**Figure 2A**).

Dimensionality-reduction analysis (UMAP) revealed distinct characteristics for each feature type (**Figures 2B–E**). Sources features (**Figure 2B**) showed clear hierarchical structures, clustering CHPs from different biological sources effectively. Medicinal property features (**Figure 2C**) moderately represented flavor and nature information but exhibited weaker category distinctions. Compatibility features (**Figure 2D**) captured synergistic relationships among CHPs but showed less defined boundaries. Efficacy features (**Figure 2E**) excelled in linking CHPs to therapeutic applications, with strong clustering patterns.

To enhance expressiveness, medicinal property, efficacy, and compatibility features were merged into a composite relational feature (**Figure 2F**), significantly improving category separation. Integrating origin features with this composite relational representation resulted in a 90-dimensional comprehensive feature vector (**Figure 2G**). UMAP projections of these integrated features displayed robust structural organization and pronounced category boundaries, validating the multi-modal feature fusion strategy for characterizing complex CHP attributes.

To evaluate dimensionality-reduction methods, t-SNE was applied as a comparison (**Figure 2H**). Results aligned with UMAP projections, showing consistent category separation and reinforcing the robustness of the multi-modal feature integration approach. This unified encoding strategy effectively integrates origin, medicinal properties, efficacy, and compatibility attributes into a single feature representation. The resulting feature vectors provide a technical foundation for data-driven characterization of CHPs, facilitating deeper exploration of TCM compatibility mechanisms and formula modeling.

### *Encoding and classification of medicinal properties and diagnostic terminologies in TCM*

The medicinal properties of CHPs are foundational for understanding their efficacy and compatibility principles. To systematically characterize relationships among these properties, we standardized key attributes, including therapeutic nature, medicinal flavor, and meridian tropism, to construct an undirected medicinal property knowledge graph. Heatmap analysis (**Figure 3A**) visualized the interaction strength among attributes, with color intensity indicating relationship strength. The optimized medicinal property network (**Figure 3B**) was refined using the ForceAtlas2 algorithm, achieving a balanced spatial node distribution and clearer logical relationships. Community detection via the Louvain method divided the network into functional modules, enhancing intra-module connectivity.

Diagnostic terminologies, a cornerstone of TCM theory, are essential for studying etiology, pathogenesis, and therapeutic principles. Using the Node2Vec algorithm, we created a 32-dimensional graph embedding diagnostic terminologies into a structured network (**Figure 3C**). UMAP dimensionality reduction (**Figure 3D**) revealed significant clustering of terminologies in two-dimensional space, providing clear bases for classification and clustering.

Hierarchical clustering of the 32-dimensional feature vectors classified diagnostic terminologies into six TCM etiological categories (**Figure 3E**). The proportional distributions of etiology, pathogenesis, treatment principles, and therapeutic methods within each category (**Figure 3F**) highlighted distinct diagnostic characteristics. Key features of these clusters are detailed in the supporting material: Metabolic and circulatory imbalances (**Figure S4**); Organ aging and immune deficiency (**Figure S5**); Lifestyle and environmental influences (**Figure S6**); Physical and mental stress, and external stimuli (**Figure S7**); Internal pathological changes and infectious factors (**Figure S8**); Dry dampness and stomach harmony (**Figure S9**). Due to limited sample size in Category 6, subsequent analyses focused on the five primary TCM etiological categories.

To further characterize these diagnostic categories, we quantified node importance using ten centrality metrics (detailed in **Figure S10**). Core diagnostic features were extracted by analyzing central feature vectors, culminating in the construction of a diagnostic feature matrix for five major TCM disease types (**Figure 3G**). This matrix systematically presents the specific distributions of etiology, pathogenesis, and therapeutic principles. It highlights core features of each disease type, offering essential data support for the quantitative and modernized analysis of TCM diagnostic terminologies.

### *Graph encoding and attention mechanisms for Chinese herbal formulas*

Using a graph encoding method with virtual nodes, we constructed a multi-dimensional relational network for CHF and systematically analyzed their complex compatibility relationships with three GNN models: Graph Transformer Network (GTN), Hypergraph Neural Network (HGNN), and Graph Attention Network (GAT) (**Figure 4A**). In the network, actual nodes represent CHPs, while virtual nodes represent medicinal properties (nature, flavor, and meridian tropism). Virtual nodes connect to CHPs with shared medicinal properties through hyperedges, forming a hierarchical hypergraph structure to model complex interactions.

The GTN employs global attention mechanisms to capture long-range interactions, making it suitable for modeling dependencies between distant nodes (**Figure 4B**). The HGNN aggregates medicinal property information via hyperedges, effectively representing high-order structural relationships (**Figure 4C**). The GAT dynamically adjusts information propagation weights through edge-specific attention, excelling in identifying key nodes and interactions within formulas (**Figure 4D**). Preliminary tests showed that increasing network layers beyond a certain threshold led to overfitting without significant performance gains. Accordingly, GTN was implemented with a single hidden layer, HGNN with two, and GAT with three, ensuring efficient attention interactions, robust interpretability, and stability.

We optimized model performance using a two-stage grid search combined with five-fold cross-validation to tune structural parameters (hidden_dim, num_heads, dropout_rate) and training parameters (learning_rate, batch_size). The optimal configurations were: GTN (hidden_dim=64, num_heads=8, dropout_rate=0.3, learning_rate=0.0007, batch_size=128), HGNN (hidden_dim=96, num_heads=4, dropout_rate=0.5, learning_rate=0.0005, batch_size=64), and GAT (hidden_dim=64, num_heads=4, dropout_rate=0.5, learning_rate=0.0001, batch_size=32). Details of the optimization process are provided in **Figure S11** and **Data S2**.

Performance was evaluated across five TCM syndrome categories—metabolic and circulatory imbalance, organ aging and immune deficiency, lifestyle and environmental factors, mental stress and external stimuli, and internal pathological changes and infections—using metrics such as AUC, precision, recall, F1 score, accuracy, and specificity (**Table S1**). AUC was a key metric to assess the model's ability to distinguish unlabeled negative samples. GTN achieved an F1 score of 0.76 for "lifestyle and environmental factors" but struggled with "mental stress and external stimuli" (AUC=0.62), highlighting its limitations in long-range interaction modeling (**Figure 4E**). HGNN excelled in "metabolic and circulatory imbalance," achieving an AUC of 0.90 on the test set (**Figure 4F**), but had lower precision (0.16) for "organ aging and immune deficiency," indicating potential false positives. GAT outperformed both models across all categories, achieving AUCs above 0.75, particularly for "mental stress and external stimuli" (AUC=0.76) and "organ aging and immune deficiency" (AUC=0.81), demonstrating superior adaptability for complex relationship modeling (**Figure 4G**).

To elucidate prediction mechanisms, feature ablation and node masking analyses were conducted (**Data S3**). Results identified the Combination feature as the most critical for AUC, highlighting its role in modeling herbal compatibility relationships. Sources and Medicinal properties ranked second, while Dosage weight had the lowest contribution (**Figure 4H**). Node masking showed that CHP nodes had the highest impact on model predictions, with AUC significantly decreasing after masking. Conversely, nodes for Therapeutic nature, Medicinal flavor, and Meridian tropism contributed less to overall model performance (**Figure 4I**). Further analysis (**Figure S12**) reinforced the importance of the Combination feature and CHP nodes.

In conclusion, the graph encoding method with virtual nodes significantly improved the modeling of multi-dimensional relationships in CHF. The GAT model, with superior classification performance and feature interpretability, emerged as the most effective for uncovering complex compatibility mechanisms. CHP nodes and Combination features were identified as core factors for CHF prediction. All research outcomes, including methods and data, are available at https://github.com/ZENGJingqi/GraphAI-for-TCM , enabling users to analyze prediction results and attention weights among CHPs. This provides robust scientific support for further studies on CHF compatibility mechanisms.

***Compatibility mechanisms in Chinese herbal formulas used for COVID-19 management***

This study analyzed 215 CHF used in the prevention and treatment of COVID-19, sourced from guidelines issued by provincial health commissions, TCM administrations, and local governments across China. Among these, 186 formulas (**Figure 5A**) were derived from regional treatment guidelines, and 29 were listed in the *Diagnosis and Treatment Protocol for Novel Coronavirus Pneumonia (Trial Version 8)* issued by China's National Health Commission. These formulas, designed for adult patients, included 167 therapeutic and 48 preventive formulas (details in **Data S4**). Using the GAT model, the distribution of formulas across five TCM syndrome categories showed that most were classified under "internal pathological changes and infection factors" (**Figure 5B**), aligning with TCM's understanding of COVID-19 etiology (*3*) and validating the model's ability to capture disease-specific mechanisms (**Data S5**).

To investigate compatibility mechanisms, we analyzed CHP usage frequency and attention weights in the model. Results (**Figure 5C**) identified Licorice (CHP01118), Honeysuckle (CHP01958), and Forsythia (CHP02298) as the most frequently used herbs, underscoring their importance in COVID-19 management. Honeysuckle and forsythia, a classic herb pair known for anti-inflammatory and detoxifying effects, exhibited high frequency and attention weights, highlighting their synergistic antiviral and anti-inflammatory roles (**Figure 5D**, **Data S6**).

Radix Astragali (CHP01717) emerged as a pivotal herb. It demonstrated high usage frequency, significant attention weight, and centrality within key herb pairs such as Radix Astragali–Perilla (CHP01717–CHP05676) and Radix Astragali–Coix Seed (CHP01717–CHP00216). These findings suggest that Radix Astragali functions as the "sovereign herb" in COVID-19 formulas, driving immune modulation and core efficacy, while "minister" or "assistant" herbs like perilla and coix seed enhance or coordinate its effects. For example, in Huashi Baidu Decoction (*17*), attention analysis revealed that Radix Astragali had the highest attention weights toward other CHPs, confirming its central role in compatibility (**Figure 5E**).

KEGG pathway enrichment analysis of Radix Astragali's compound-target interactions identified three primary mechanisms: the neuroactive ligand-receptor interaction pathway (*18*), the MAPK signaling pathway (*19*), and the calcium signaling pathway (*20*) (**Figure 5F**). These pathways are associated with inflammation regulation, immune modulation, and antiviral effects, further emphasizing its significance in COVID-19 management.

Compound analysis revealed that Radix Astragali's pharmacological effects are primarily driven by terpenoids, alkaloids, and shikimates and phenylpropanoids (**Figure 5G**). Filtering compound-target pairs with binding affinities >8.0, we constructed a component-compound-target-KEGG pathway network for Radix Astragali (**Figure 5H**), showing that high-affinity compounds are integral to key pathways (**Data S7**).

Combining compound drug-likeness screening, we identified potential active molecules in Radix Astragali with high drug development potential (**Figure 5I**). These molecules, characterized by strong binding affinities and broad target coverage, present promising opportunities for drug discovery.

In conclusion, this study utilized a graph neural network to analyze CHP attention weights and identified Radix Astragali as a core herb in COVID-19 formulas. These findings elucidate its key role in compatibility mechanisms and provide a scientific foundation and data support for pharmacological research and drug development.

## Discussion

This study utilized GraphAI technologies integrated with TCM theories to quantitatively analyze the complex compatibility mechanisms in CHF. It addresses critical challenges in TCM research, including the quantification of herbal synergy, the interpretability of models, and the scarcity of high-quality data resources. By employing the GAT model, we uncovered synergistic effects among herbs, quantified the contributions of core CHP, and provided new pathways for modern TCM research and drug discovery.

Unlike prior studies that focus on formula recommendations (*21–24*), our research emphasizes the elucidation of compatibility mechanisms in clinically validated formulas rather than designing new ones. By introducing virtual nodes for medicinal properties and employing multi-dimensional feature representations, our model achieves a high level of interpretability. Outputs from the GAT model are transferable to small-sample problems in specific diseases, overcoming the "black-box" limitations of deep learning. This novel approach supports the optimization of existing formulas and lays a foundation for material basis research and new drug development.

The "sovereign-minister-assistant-courier" framework is a cornerstone of TCM compatibility theory. For the first time, this study quantitatively evaluated CHP contributions using GraphAI and attention mechanisms. Results revealed that core herbs such as Radix Astragali dominate key herb pairs, with KEGG pathway enrichment confirming its immunomodulatory role. Radix Astragali's pivotal function in regulating inflammation and immune responses underscores its role as the "sovereign herb" in COVID-19 treatment formulas, providing a clear basis for further formula optimization and material basis exploration.

Long COVID, a complex syndrome characterized by immune dysregulation, viral persistence, and endothelial dysfunction (*25*), may benefit from TCM interventions centered on Radix Astragali. Through mechanisms such as the MAPK signaling, calcium signaling, and neuroactive ligand-receptor interaction pathways, Radix Astragali demonstrates significant immunomodulatory and anti-inflammatory effects. For instance, a preventive formula recommended in Shanghai, composed of food-medicinal herbs (Radix Astragali 15g, Reed Rhizome 15g, Lotus Leaf 6g, Patchouli 6g, Perilla Leaf 6g, Mint 3g), offers a safe and practical option for improving immune function and managing long COVID.

While compound-target analysis provides valuable insights into drug mechanisms, TCM's efficacy extends beyond small-molecule actions. This study emphasized the holistic synergistic effects of herbal combinations alongside mechanistic analyses, highlighting the potential for multi-target synergy to produce enhanced pharmacological effects. By balancing holistic efficacy with mechanistic insights, this dual-focus paradigm offers a comprehensive understanding of TCM.

A key strength of this study is the open accessibility of its data and code resources. These high-quality resources foster academic research, interdisciplinary collaboration, and technological innovation. The proposed framework is adaptable to other traditional medical systems, such as Uyghur medicine, enabling exploration of unique theories and strategies. This approach facilitates cross-cultural comparisons, elucidating commonalities and differences among medical systems while offering new perspectives for global healthcare services.

In conclusion, this study introduces an innovative method to quantify TCM compatibility mechanisms, validating the modern potential of TCM theories by highlighting Radix Astragali's central role in COVID-19 treatment formulas. Future efforts will expand this framework, develop multi-expert systems, and strengthen the scientific foundations of TCM theories and practices. By optimizing healthcare services and advancing drug development, this work contributes to the global modernization of traditional medicine.

## Materials and Methods

### *TCM terminology module*

The TCM terminology module forms the semantic foundation of the TCM-MKG. Systematic categorization and standardization ensure semantic consistency across modules. This module is based on the WHO International Standard Terminologies on Traditional Medicine (2022 edition**)**, comprising 1,810 core TCM terms (ISBN: 9789240042322), including key concepts such as etiology, pathogenesis, therapeutic principles, and treatments. These terminologies provide theoretical support for cross-module data interconnectivity. The standard, published by the **World** Health Organization (WHO), is accessible at: https://www.who.int/publications/i/item/9789240042322.

### *Chinese patent medicine module*

The CPM module integrates data on 8,977 Chinese patent medicines, including formula composition, dosage forms, efficacy, indications, and links to modern medical diseases. Primary data sources include the Chinese Pharmacopoeia (2020 edition) and the China Medicine Information Query Platform (www.dayi.org.cn). Additionally, CPM indications are mapped to ICD-11 codes using the ICD-11 coding tool (https://icd11.pumch.cn) to establish a bridge between traditional Chinese and modern Western medical systems.

### *Chinese herbal pieces module*

The CHP module documents 6,207 Chinese herbal pieces, encompassing essential information such as medicinal properties, biological sources, and chemical compositions. This module supports systematic investigations into the mechanisms of action and chemical characteristics of CHPs. Key data sources include the Chinese Pharmacopoeia (2020 edition), the China Medicine Information Query Platform (www.dayi.org.cn), and the NCBI Taxonomy (https://www.ncbi.nlm.nih.gov/taxonomy) (*26*), which standardizes Latin species names.

Chemical composition data were integrated from authoritative databases and lab-curated sources, including: TCMID V2.0 (http://47.100.169.139/tcmid) (*27*); TCMbank V1.0; (https://tcmbank.cn) (*28*); SymMap V2.0 (http://www.symmap.org) (*29*); HIT V2.0 (http://hit2.badd-cao.net) (*30*); HERB V2.0 (http://47.92.70.12) (*31*); ETCM V2.0 (http://www.tcmip.cn/ETCM2) (*32*); TCMSP V1.0 (https://www.tcmsp-e.com/) (*33*).

### *Natural products module*

The NP module documents 4,810 natural products, providing comprehensive information on their biological sources, chemical compositions, and related properties. This module forms a systematic foundation for studying natural products in TCM, supporting research on pharmacological effects and molecular mechanisms. Data primarily originate from NCBI Taxonomy (https://www.ncbi.nlm.nih.gov/taxonomy) (*26*), ensuring precise classification of natural products (e.g., family, genus, species). Shared chemical composition data link the NP and CHP modules. Additional information was sourced from the CMAUP database (https://www.bidd.group/CMAUP/index.html) (*34*). For natural products identified as CHPs, mapping is achieved through CHP-NP associations. For products recorded directly as natural substances (e.g., plants or fungi identified by Latin names), chemical composition data were compiled independently.

### *Chemical compounds module*

The chemical compounds module systematically records information on 123,647 chemical compounds, including structural data, target associations, and cross-database mappings. This module serves as a robust foundation for investigating TCM molecular mechanisms and pharmacological activities. All chemical compounds were standardized using RDKit (https://www.rdkit.org/), with InChIKey as the unique identifier to ensure data consistency and traceability. Core data include SMILES, InChI, and Molecular_formula, while molecular descriptors (e.g., molecular weight, logP, topological polar surface area) generated by RDKit support structural and efficacy analyses.

To classify chemical structures, NPClassifier V1.5 (https://npclassifier.ucsd.edu/) (*35*) was used for three-tier classification (Pathway, Superclass, and Class) and for annotating IsGlycoside information to identify glycoside compounds. This classification highlights biological characteristics and pharmacological activities of the compounds. Chemical compound-target association data were integrated from multiple authoritative sources: DrugBank V5.1.12 (https://go.drugbank.com) (*36*). TTD (https://db.idrblab.net/ttd/) (*37*). STITCH V5.0 (http://stitch.embl.de) (*38*). BindingDB (https://www.bindingdb.org) (*39*). All targets were standardized using EntrezID and restricted to human-specific targets to ensure accuracy. Beyond chemical compounds derived from CHPs and natural products, the module also includes all compounds from DrugBank and TTD, as well as compound-target associations from STITCH with a confidence score above 700. This ensures the availability of high-quality training data for target prediction.

### *Targets module*

The targets module in the TCM-MKG includes 20,053 standardized human targets and 6,879,008 protein-protein interaction (PPI) records, providing a systematic foundation for TCM target research and network biology analysis. Target data were standardized using multiple identifiers, including EntrezID, UniProtID, GeneSymbol, and ENSGID, ensuring compatibility and traceability. Target symbols and sequence information were derived from Ensembl GRCh37.p13 (http://grch37.ensembl.org/Homo_sapiens/Info/Index) (*40*) and UniProt Release 2024_02 (https://www.uniprot.org/) (*41*). Each target entry includes multiple standard identifiers and detailed protein sequence data.

Protein-protein interaction data were integrated from the following authoritative resources: STRING V12.0 (https://cn.string-db.org/) (*42*). SIGNOR V3.0 (https://signor.uniroma2.it/) (*43*). MINT (https://mint.bio.uniroma2.it/) (*44*). IntAct (https://www.ebi.ac.uk/intact/home) (*45*). BioGRID V4.4.233 (https://thebiogrid.org/) (*46*). These databases provide extensive information on direct PPIs and causal relationships, facilitating in-depth analysis of TCM-related targets.

### *Disease module*

The disease module in the TCM-MKG integrates core data from the 2024 Chinese and English versions of ICD-11, extending it through standardized mappings to multiple authoritative databases, including: NCBI Unified Medical Language System (UMLS)

(https://uts.nlm.nih.gov/uts/umls). Medical Subject Headings (MeSH) (https://uts.nlm.nih.gov/uts/umls) (*47*). Disease Ontology (DOID) (https://disease-ontology.org/do) (*48*). This module includes multi-layered mappings between ICD-11 codes and corresponding UMLS Concept Unique Identifiers (CUIs), MeSH identifiers, and DOIDs, covering 12 semantic types such as diseases, syndromes, neoplastic processes, signs or symptoms, and injuries or poisoning. It comprises:65,758 CUI mapping records. 6,853 MeSH mapping records. 4,803 DOID mapping records.

These mappings enable standardized disease data management and cross-platform queries. Disease-target association data were integrated from multiple authoritative databases, including: DisGeNET V7.0 (https://www.disgenet.org/) (*49*), providing 575,153 CUI-classified disease-target associations. Comparative Toxicogenomics Database (CTD) (https://ctdbase.org/) (*50*), contributing 24,097,586 MeSH-classified disease-target associations. Diseases V2.0 (https://diseases.jensenlab.org/) (*51*), offering 6,522,131 DOID-classified disease-target associations. These comprehensive datasets support the integration of traditional TCM theories with modern biomedical frameworks, enabling cross-disciplinary research and advanced network analysis.

### *Encoding medicinal properties of Chinese herbal pieces and network construction*

This study systematically encoded and analyzed the medicinal properties of CHP using standardized processing and graph intelligence methods. The medicinal properties included therapeutic nature, medicinal flavor, and meridian tropism, which were organized into a hierarchical classification to build a medicinal property knowledge graph. To quantify the relationships among these attributes, we constructed an undirected network model, where nature, flavor, and meridian tropism were represented as nodes, and the edge weights reflected the strength of their associations.

The initial network layout was based on the logical framework of medicinal property theory, subsequently optimized using the ForceAtlas2 algorithm for enhanced visual clarity. Key parameters included an edge weight influence of 1.0, jitter tolerance of 0.1, Barnes-Hut optimization setting of 1.2, and the activation of a strong gravity mode (gravity value: 5.0). To identify potential functional modules within the network, the Louvain method was applied for community detection, grouping nodes into distinct communities based on edge weights, with modular distribution visually represented through color coding.

### *Encoding and clustering of TCM diagnostic terms*

To analyze the multi-dimensional relationships among TCM diagnostic terms, we applied the Node2Vec algorithm, embedding diagnostic features such as etiology, pathogenesis, treatment principles, and therapeutic methods into a 32-dimensional vector space. A graph model was constructed with formulas and their diagnostic features as nodes, and their relationships were represented as edge weights. The high-dimensional feature embeddings were reduced to a two-dimensional space using UMAP to reveal potential relationships among diagnostic terms.

Hierarchical clustering was subsequently performed on the 32-dimensional feature vectors to classify the diagnostic terms into six categories, with the proportion of etiology, pathogenesis, treatment principles, and therapeutic methods calculated for each category. Due to the small sample size of category six, analysis focused on the primary five categories. Core features for each category were extracted using centroid vectors to support naming.

### *Graph encoding of Chinese herbal formulas and message passing mechanisms*

To quantify the complex relationships between CHP and medicinal properties within CHF, we designed a graph encoding method using virtual nodes. This method integrates core concepts of TCM theory with modern GNN techniques for a hierarchical and quantitative analysis of formula compatibility. The graph structure consisted of real nodes and virtual nodes. Real nodes represented CHP, with attributes including 91 feature variables (90 describing characteristics such as efficacy and origin, and one for dosage, set to zero if unknown). Virtual nodes represented medicinal property categories, including "Therapeutic nature," "Medicinal flavor," and "Meridian tropism," with initial attributes calculated as the weighted mean of all connected real node attributes. Edge attributes quantified the relationships between CHP and medicinal properties using a two-dimensional encoding.

To represent the complex relationships more comprehensively within formulas, virtual node attributes dynamically updated based on the attributes of connected real nodes, calculated as the weighted mean of all linked node attributes. Edge attributes between virtual nodes were determined by averaging the attributes of all connected real-node edges, reflecting potential relationships among medicinal property categories. This design, through virtual nodes connecting multiple real nodes, established a hierarchical hypergraph structure for CHF, facilitating the discovery of complementary and interactive relationships among CHP.

After constructing the graph structure, we implemented three GNN models to capture the multi-dimensional relationships within CHF: GTN: Captures long-range interactions using a global attention mechanism between nodes. A single-layer hidden design balances complexity and computational efficiency. HGNN: Leverages hyperedges to organize high-order structural relationships between virtual and real nodes, using two hidden layers to enhance information aggregation. GAT: Dynamically adjusts message-passing weights through edge attention mechanisms, focusing on key interactions among CHP. A three-layer hidden structure captures multi-level node interactions. This design ensures that every node in the model interacts attentively with others, achieving high interpretability while improving modeling efficiency and stability.

### *Data selection and grouping for graph neural network models*

This study utilized a dataset containing 6,080 CHF, categorized into five major disease types: metabolic and circulatory imbalance, organ aging and immune deficiency, lifestyle and environmental impacts, psychological stress and external stimuli, and internal pathological changes with infectious factors. The dataset was split into training (4,256 samples), validation (1,222 samples), and test (602 samples) sets in a 7:2:1 ratio. The positive sample proportions for the five disease types in the training, validation, and test sets were as follows: metabolic and circulatory imbalance (6.34%, 6.87%, 4.32%), organ aging and immune deficiency (6.41%, 6.79%, 6.64%), lifestyle and environmental impacts (57.31%, 57.77%, 57.81%), psychological stress and external stimuli (10.10%, 9.17%, 9.97%), and internal pathological changes with infectious factors (30.45%, 30.03%, 32.56%). To comprehensively evaluate model performance across the five disease types, multiple metrics were employed, including precision, recall, F1 score, area under the curve (AUC), accuracy, and specificity.

### *Hyperparameter optimization*

To optimize the performance of the three GNN models—GTN, HGNN, and GAT—a two-stage grid search strategy was employed, combined with 5-fold cross-validation using the full dataset of 6,080 samples. In the first stage, model structural parameters were tuned, including hidden layer dimensions (hidden_dim: 32, 64, 96, 128), number of attention heads (num_heads: 2, 4, 8, 16), and dropout rates (dropout_rate: 0.1, 0.3, 0.4, 0.5). The average loss on the validation set was used to assess the performance of each parameter combination, and the configuration with the lowest loss was selected as the optimal structural parameter set. In the second stage, training parameters such as learning rate (learning_rate: 0.0001, 0.0003, 0.0005, 0.0007) and batch size (batch_size: 16, 32, 64, 128) were optimized. These combinations were evaluated based on the average loss on the validation set, with the lowest loss configuration chosen as the optimal training setup. To ensure stability, all parameter searches employed 5-fold cross-validation, using the mean validation loss as the evaluation metric. Detailed results of the optimization process and raw data are available in the supporting materials (**Figure S11** and **Data S2**).

### *Model interpretability analysis*

To assess the contribution of features and nodes in the models, this study employed the feature nullification method for interpretability analysis of the GAT model. The feature nullification method involves setting specific feature values to zero (nullification) and observing changes in model performance (AUC), thereby quantifying the importance of each feature to prediction outcomes. For node contribution analysis, all features of specific nodes (e.g., dosage, medicinal properties) were nullified one at a time, and the AUC before and after nullification was compared to evaluate the node's impact on prediction results. All analyses were conducted on the test set to avoid overfitting and ensure the representativeness and robustness of interpretability results. Raw data from these analyses are available in **Data S3**.

**Acknowledgements**

This work benefited from the integration of data from numerous open-access and authoritative databases. We acknowledge the valuable contributions of resources such as DrugBank, BindingDB, BioGRID, DisGeNET, and many others. These datasets provided essential insights into TCM, modern drug chemistry, genetics, diseases, and related fields, forming the foundation for the TCM Multi-dimensional Knowledge Graph (TCM-MKG) used in this study. Furthermore, we utilized the PSICHIC model (https://github.com/huankoh/PSICHIC ) to analyze the binding interactions between components and targets. Full citations for these resources are included in the manuscript.

**Authors' contributions**

Conceptualization: Xiaobin Jia; Methodology: Jingqi Zeng; Investigation: Jingqi Zeng; Data curation: Jingqi Zeng; Formal analysis: Jingqi Zeng; Software: Jingqi Zeng; Validation: Jingqi Zeng; Visualization: Jingqi Zeng; Funding acquisition: Xiaobin Jia; Project administration: Xiaobin Jia; Supervision: Xiaobin Jia; Writing – original draft: Jingqi Zeng; Writing – review & editing: Jingqi Zeng, Xiaobin Jia.

**Competing interests**

The authors declare that they have no competing interests.

**Funding**

This research was supported by the National Natural Science Foundation of China (Grant No. 82230117, Xiaobin Jia). The funders of the study had no role in study design, data collection, data analysis, data interpretation, or writing of the report.

**Data and materials availability**

All data and code used in this study are openly available. The TCM multi-dimensional knowledge graph (TCM-MKG) is accessible at https://zenodo.org/records/13763953 , and the graph artificial intelligence model implementation is available at https://github.com/ZENGJingqi/GraphAI-for-TCM .

**Supplementary Materials**

Supplementary Text

Figs. S1 to S12

Tables S1

Data S1 to S7

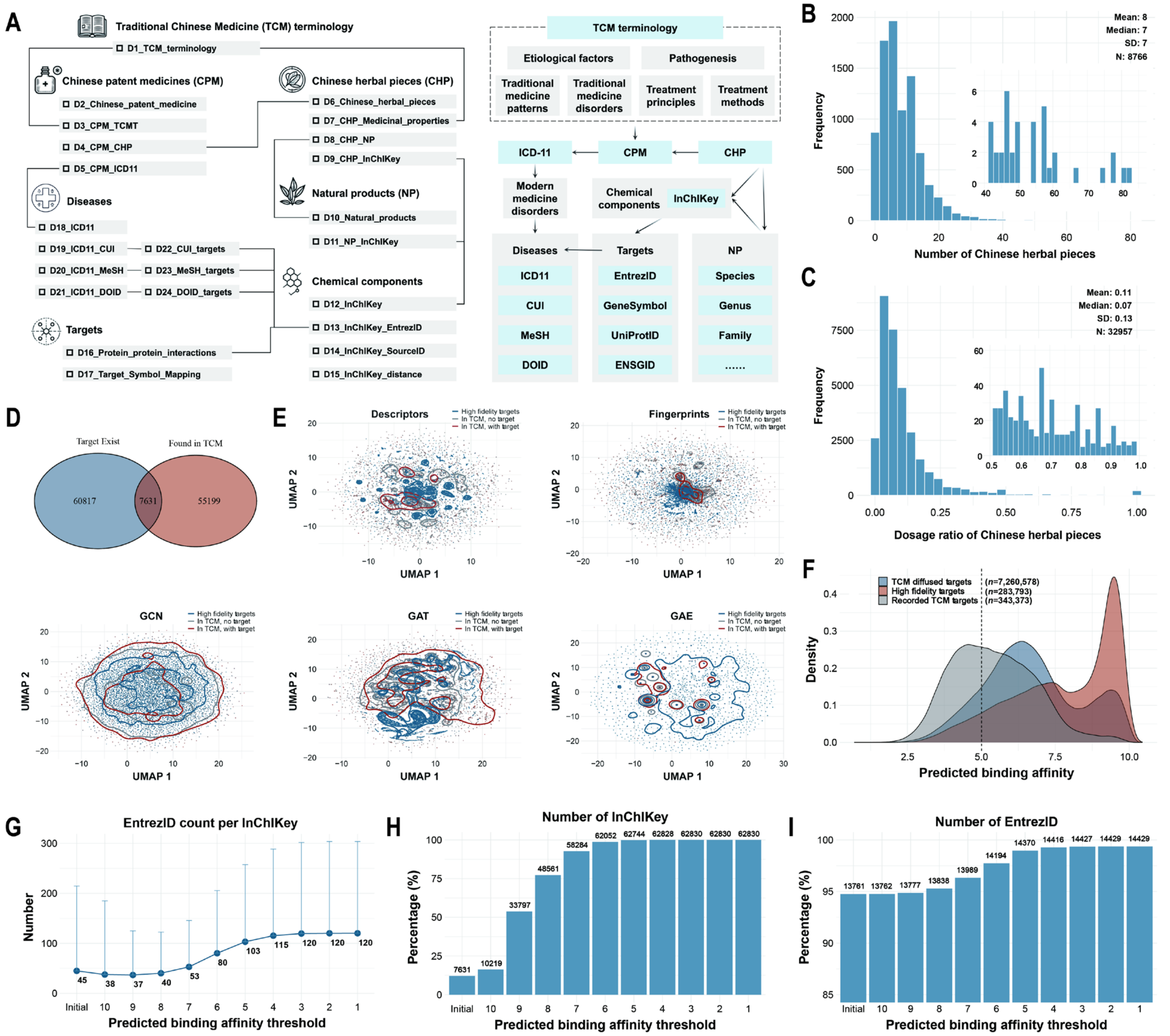

**Fig. 1. Data structure and component-target analysis in the TCM-MKG.** (A) Overview of the TCM-MKG database structure, including TCM terminology, Chinese patent medicines (CPM), Chinese herbal pieces (CHP), natural products (NP), chemical components, targets, and diseases. (B) Distribution of the number of Chinese herbal pieces per CPM. (C) Dosage ratio distribution of Chinese herbal pieces in CPMs. (D) Presence of compound targets in the TCM database. (E) UMAP visualization of molecular descriptors, Morgan fingerprints, and graph-based models: graph convolutional network (GCN), graph attention network (GAT), and graph autoencoder (GAE). (F) Binding affinity distribution of three datasets: diffused TCM targets (7,260,578 samples), high-fidelity targets (283,793 samples), and recorded TCM targets (343,373 samples), with a random control of 10,000 non-TCM targets. (G) Number of EntrezIDs per InChIKey across binding affinity thresholds. (H) InChIKey coverage across binding affinity thresholds. (I) EntrezID coverage across binding affinity thresholds.

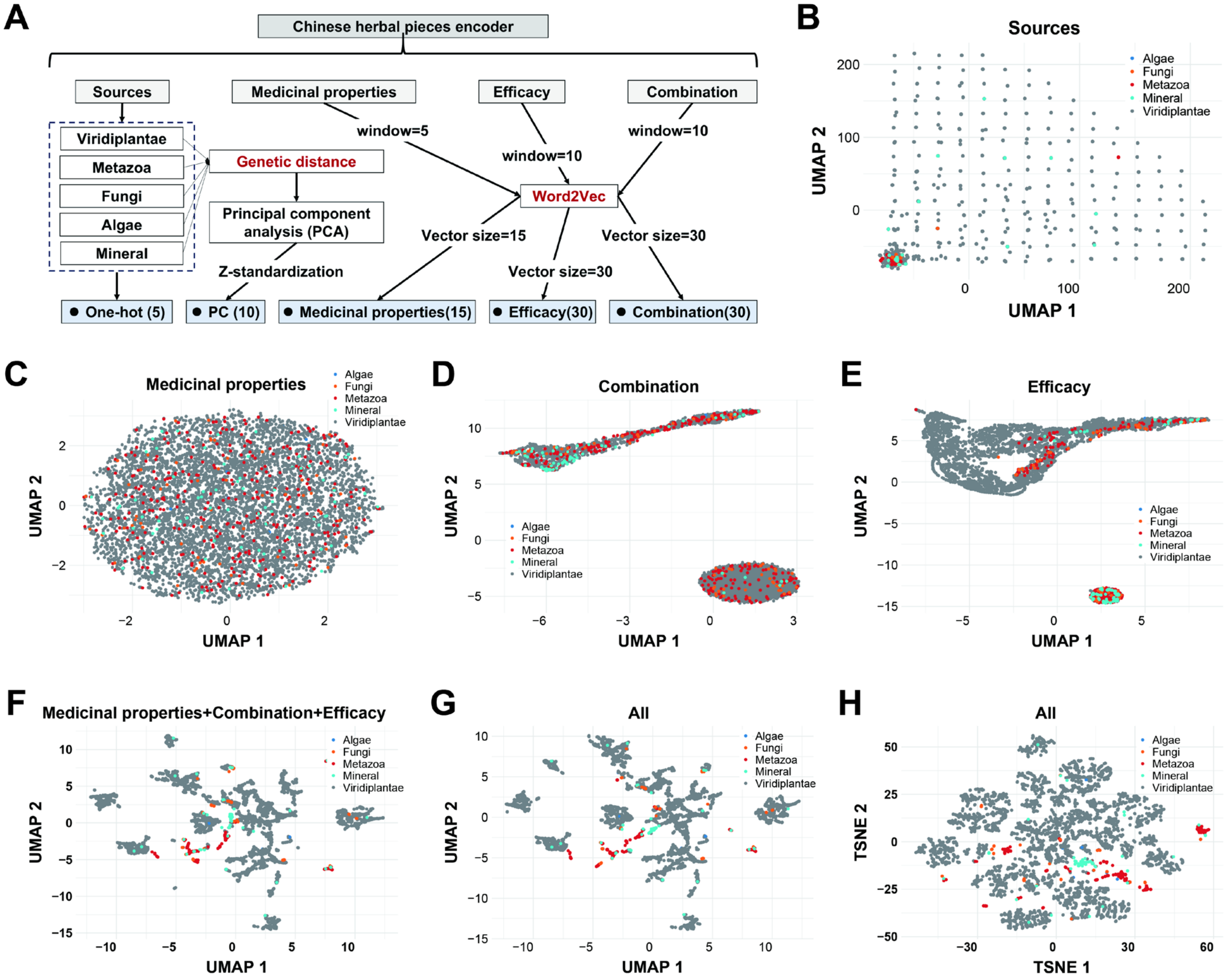

**Fig. 2. Multi-feature encoding and dimensionality reduction analysis of Chinese herbal pieces.** (A) Workflow of multi-feature encoding for Chinese herbal pieces. Features include sources (encoded with one-hot and principal component analysis, PCA), medicinal properties (vectorized using Word2Vec with window size = 5 and vector size = 15), efficacy (vectorized using Word2Vec with window size = 10 and vector size = 30), and combination relationships (vectorized using Word2Vec with window size = 10 and vector size = 30). Z-standardization was applied where necessary. Encoded feature dimensions are labeled as: sources one-hot (5), sources PCA (10), medicinal properties (15), efficacy (30), and combination (30). (B) UMAP projection of source features. (C–E) UMAP projections of features representing medicinal properties, combination relationships, and efficacy, respectively. (F) UMAP projection of combined features: medicinal properties, combination, and efficacy. (G) UMAP projection of all combined features, integrating sources, medicinal properties, efficacy, and combination. (H) t-SNE projection of all combined features.

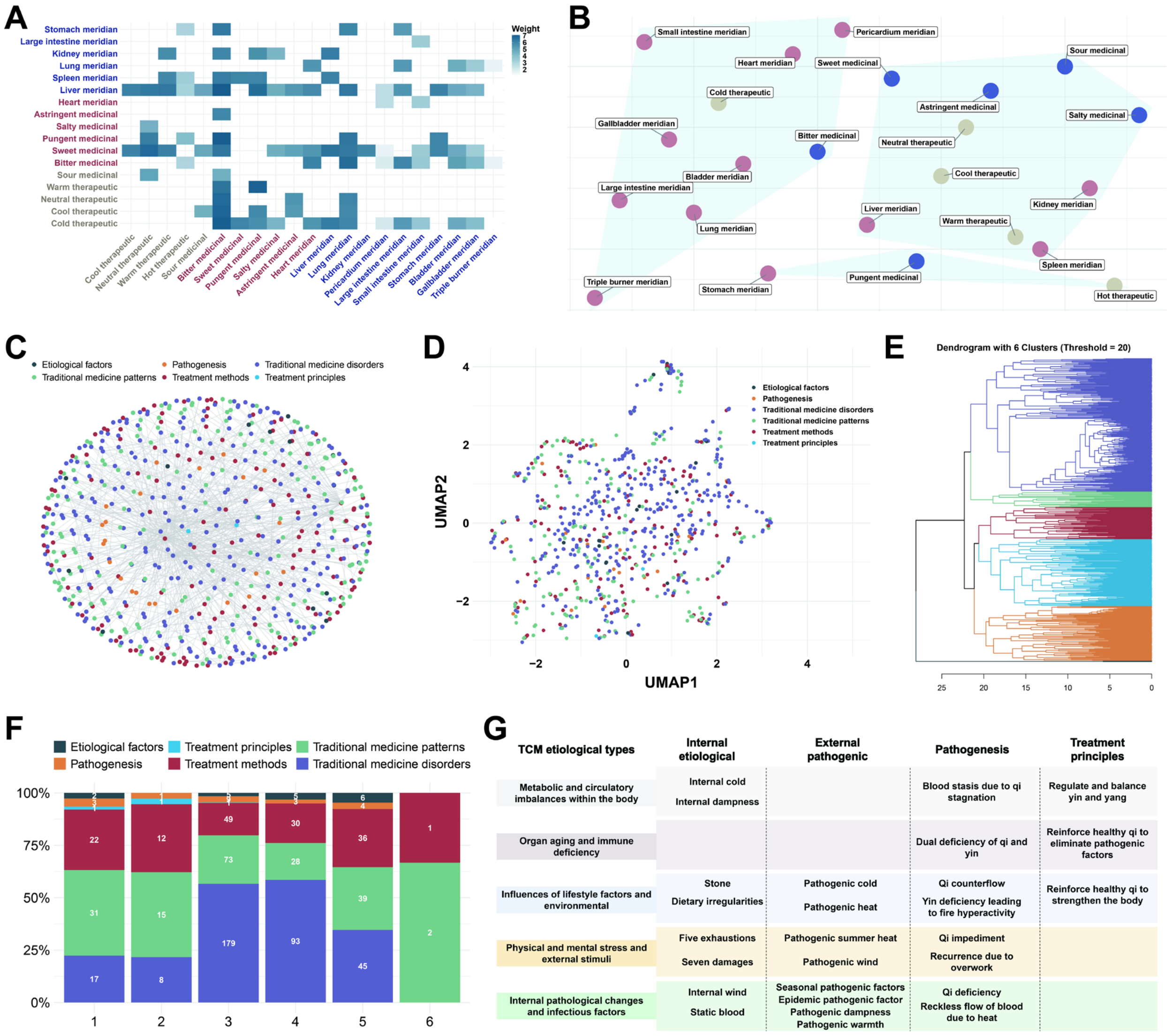

**Fig. 3. Analysis of medicinal properties and TCM diagnostic terminology.** (A) Heatmap showing the association strengths between medicinal properties of Chinese herbal pieces, categorized into therapeutic nature, medicinal flavor, and meridian tropism. (B) Network distribution of medicinal properties, visualized with community detection and optimized layout, highlighting functional modules and interrelationships. (C) Network structure of TCM diagnostic terminologies, encompassing etiological factors, pathogenesis, traditional medicine patterns, treatment methods, and treatment principles, encoded into 32-dimensional vectors using Node2Vec. (D) UMAP projection of TCM diagnostic terminologies. (E) Hierarchical clustering dendrogram of TCM diagnostic terminologies, classified into six major groups. (F) Proportional distribution of diagnostic terminologies across six TCM etiological categories. (G) Diagnostic characteristics of five major TCM etiological types, detailing internal etiological factors, external pathogenic factors, pathogenesis, and treatment principles. The sixth category is excluded due to insufficient data.

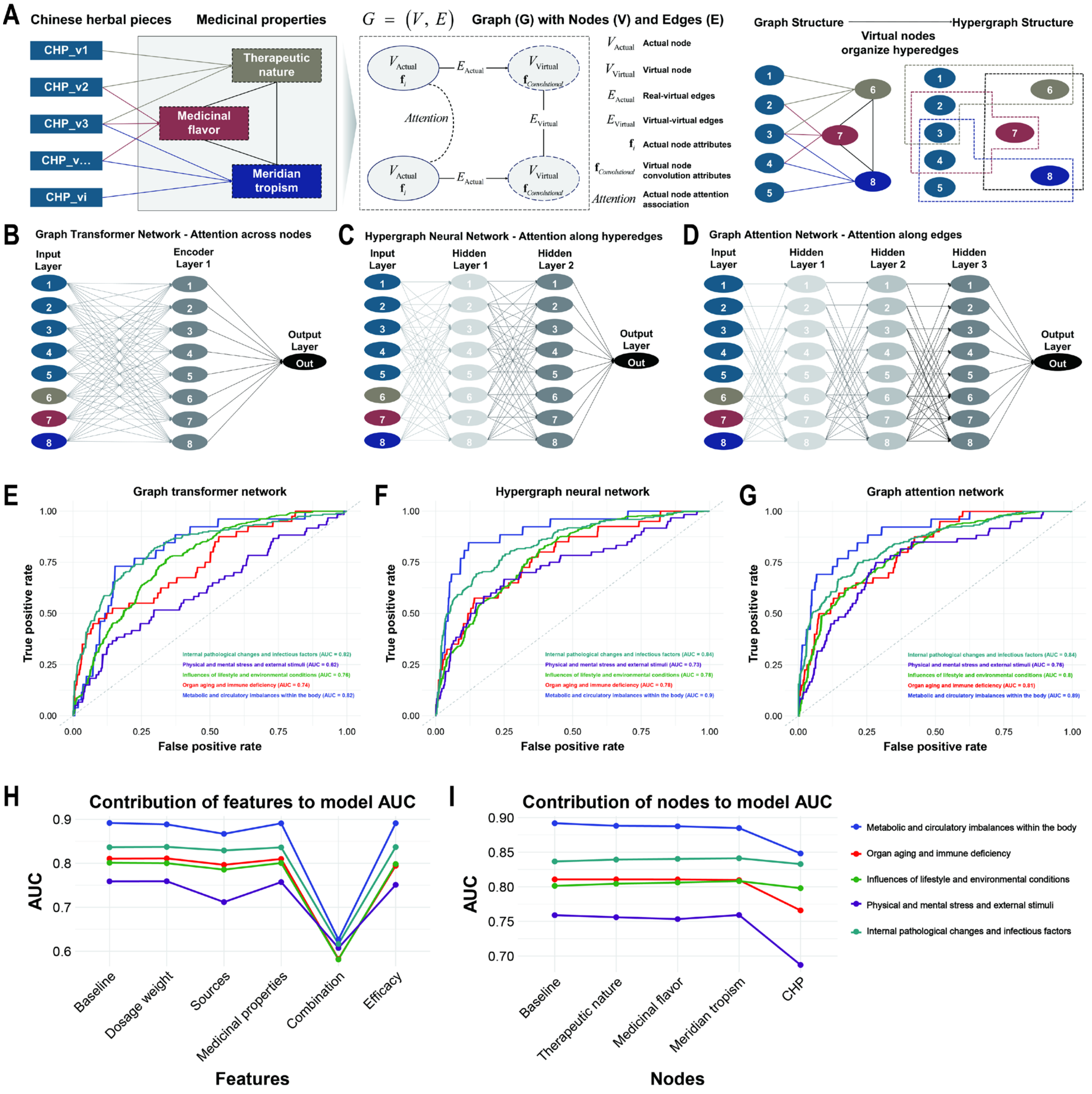

**Fig. 4. Graph encoding and attention mechanisms for Chinese herbal formulas.** (A) Graph structure encoding the relationships between Chinese herbal pieces and their medicinal properties, represented by virtual nodes for therapeutic nature, medicinal flavor, and meridian tropism. Virtual nodes enable transformation into hypergraph structures for advanced modeling. (B–D) Architectures of graph models: (B) graph transformer network, (C) hypergraph neural network, and (D) graph attention network. (E–G) AUC performance curves for (E) graph transformer network, (F) hypergraph neural network, and (G) graph attention network, across five major TCM etiological categories. (H–I) Contributions to model AUC evaluated using the zeroing method: (H) feature-level contributions and (I) node-level contributions, both based on the graph attention network.

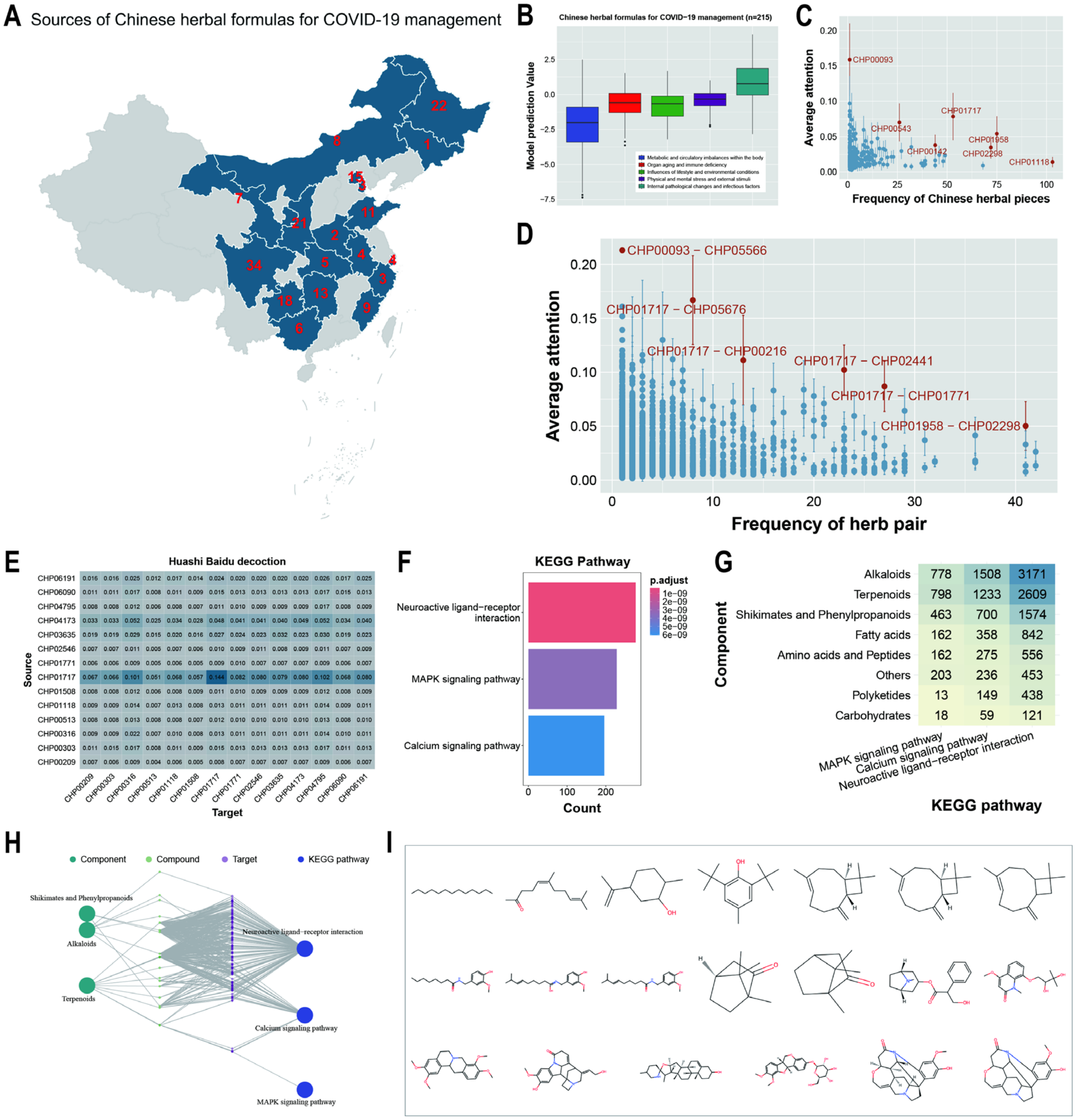

**Fig. 5. Analysis of compatibility mechanisms in Chinese herbal formulas for COVID-19 management.** (A) Geographic distribution of 215 Chinese herbal formulas across China, excluding 29 nationwide formulas from the National Health Commission. (B) Model-predicted outcomes of the formulas across five major TCM etiological categories. (C–D) Frequency and average attention weights of Chinese herbal pieces and herb pairs in the formulas, with key elements highlighted based on high frequency or significant attention weights. (E) Attention weights of Chinese herbal pieces in the Huashi Baidu Decoction, with darker heatmap colors indicating higher weights and numerical values displayed. (F) Top three KEGG pathways enriched for Radix Astragali (CHP01717) targets. (G) Contributions of Radix Astragali component to KEGG pathways, with numbers indicating compound-target pairs per pathway. (H) Component-compound-target-KEGG pathway network for Radix Astragali. (I) Key bioactive compounds of Radix Astragali with predicted binding affinities >8.0 to targets in enriched KEGG pathways.

# Supplementary Materials

## Graph Neural Networks for Quantifying Compatibility Mechanisms in Traditional Chinese Medicine

Jingqi Zeng, Xiaobin Jia

Corresponding author: jiaxiaobin2015@163.com

**The PDF file includes:**

Supplementary Text
Figs. S1 to S12
Tables S1

**Other Supplementary Materials for this manuscript include the following:**

Data S1 to S7

## Supplementary Text

### *Impact of feature representation methods and neighbor distances on predicted binding affinity*

In predicting binding affinities for TCM compound-target interactions, both feature representation methods and neighbor distances exhibited clear trends. Shorter neighbor distances consistently correlated with higher predicted binding affinities across all feature representation methods, emphasizing that compounds with similar features are more likely to bind to high-affinity targets.

Traditional feature representations, such as molecular descriptors (**Figure S1A**) and molecular fingerprints (**Figure S1B**), demonstrated moderate efficacy. Compounds at shorter neighbor distances were predominantly distributed in the medium binding affinity range (5.0–7.5), indicating that these methods effectively capture local structural similarities. However, their performance in the high-affinity range (8.0–10.0) was limited, reflecting challenges in representing complex molecular features.

Graph neural network (GNN)-based representations showed significant improvements in predicting binding affinities. The Graph Autoencoder **(**GAE, **Figure S1C)** and Graph Convolutional Network **(**GCN, **Figure S1E)** performed robustly across both moderate (5.0–7.5) and high-affinity (8.0–10.0) ranges. These methods excel at extracting nuanced molecular graph features, capturing local similarities and molecular properties linked to high-affinity targets. At shorter neighbor distances, GAE and GCN demonstrated high concentrations of predictions in the high-affinity range. As distances increased, their ability to predict high-affinity interactions diminished, underscoring their strength in identifying high-quality similarities within short ranges.

The Graph Attention Network **(**GAT, **Figure S1D)** showed intermediate performance, bridging the gap between traditional methods and GAE/GCN. While GAT predicted higher binding affinities at shorter neighbor distances, its predictions were primarily concentrated in the moderate affinity range (5.0–7.5). Its capacity to identify high-affinity targets was less pronounced, suggesting that, while GAT's multi-head attention mechanism enhances the representation of compound-target similarities, it falls short compared to GAE and GCN in capturing complex molecular features associated with high-affinity interactions.

To quantify the performance of each method, the average binding affinities within short neighbor distances were calculated (**Figure S1F**). Results showed that GAE and GCN achieved significantly higher average affinities compared to molecular descriptors, molecular fingerprints, and GAT. These findings confirm the superiority of GAE and GCN embeddings for high-affinity target prediction, offering a more robust framework for exploring TCM compound-target interactions.

### *Influence of molecular weight and target sequence length on binding affinity prediction*

This study analyzed binding affinity patterns for "High fidelity targets," "Recorded TCM targets," and "TCM diffused targets" based on molecular weight and target sequence length. The results highlight significant differences in binding affinity distributions, reflecting the impact of data generation methods and target selection strategies.

The "High fidelity targets" dataset **(Figure S2A)** showed a concentrated distribution of high binding affinities (mean affinity > 7.5) for compounds with molecular weights between 350 and 1100 Da and target sequence lengths under 1000. This pattern aligns with target-driven screening processes, such as rational design or target-specific optimization, which prioritize compound-target pairs with high compatibility. The dataset prominently features affinities in regions combining moderate molecular weights with short sequence targets, emphasizing a focused selection process.

In contrast, the "Recorded TCM targets" dataset **(Figure S2B)** displayed a more dispersed affinity distribution. High-affinity regions (mean affinity > 6.5) were concentrated around compounds with molecular weights of 354–454 Da, interacting with targets of various lengths. This dispersion likely results from compound-driven methods, such as virtual screening or structural similarity-based approaches, which expand target associations but increase the risk of false positives. The complex chemical structures and multi-target nature of TCM compounds further contribute to the wider distribution of binding affinities.

The "TCM diffused targets" dataset **(Figure S2C)** exhibited a distribution pattern similar to "High fidelity targets." High binding affinities (mean affinity > 7.0) were concentrated in the same molecular weight (350–1100 Da) and sequence length (<1000) regions. This indicates that the neighbor-diffusion strategy effectively inherits the characteristics of "High fidelity targets" when expanding compound-target associations. By leveraging structural similarity among compounds, the neighbor-diffusion method infers nearby compound-target associations, ensuring that the binding affinity distribution aligns closely with that of "High fidelity targets." This approach balances expanded data coverage with prediction reliability.

In summary, the binding affinity distribution of "High fidelity targets" reflects the focused nature of target-driven design, while "Recorded TCM targets" exhibit greater dispersion due to compound-centric expansion methods. The "TCM diffused targets" demonstrate a concentrated distribution comparable to "High fidelity targets," validating the reliability of the neighbor-diffusion method. These findings confirm that the neighbor-diffusion strategy effectively expands TCM compound-target data while maintaining high confidence in predictions, providing robust support for elucidating TCM molecular mechanisms.

### *Feature encoding of sources attributes for Chinese herbal pieces*

Sources attributes are crucial features of CHPs, reflecting their biological sources and evolutionary relationships. Using phylogenetic tree analysis and PCA, this study constructed a 15-dimensional feature vector integrating biological classification and evolutionary information. The workflow and results are detailed in **Figure S3**. As illustrated in **Figure S3A**, the majority of CHPs originate from the Viridiplantae, followed by Metazoa, minerals, Fungi, and Algae, highlighting the proportional contributions of various biological sources in TCM. TaxID information from the NCBI Taxonomy database was used to construct a phylogenetic tree based on taxonomic positions. The resulting structures for minerals, plants, animals, fungi, and algae are shown in **Figures S3B–F**, respectively. Mineral drugs (**Figure S3B**) were encoded as independent nodes due to their non-living nature, while other groups exhibited complex evolutionary branching.

PCA on the phylogenetic tree distance matrix quantified the sources features. The top 10 principal components captured over 80% of the variance for plants, animals, and fungi (**Figure S3G**). Due to the limited sample size of algae (seven species), a maximum of seven principal components was extracted, and the remaining three features were supplemented with dummy variables, similar to the encoding of minerals. This ensured a consistent 15-dimensional vector while preserving algae-specific properties. Additionally, the five primary sources categories (plants, animals, fungi, algae, and minerals) were encoded with binary dummy variables, providing explicit categorical information. These features enhanced the adaptability of models to diverse CHP sources.

In summary, the sources feature encoding produced a 15-dimensional vector composed of 10 principal components and 5 dummy variables, effectively capturing biological classification and evolutionary relationships. This encoding establishes a robust foundation for systematic CHP research.

### *Identification of core nodes in the TCM diagnostic-therapeutic terminology network*

To identify key nodes in the TCM diagnostic-therapeutic terminology network, node importance was quantitatively assessed using ten centrality metrics, including degree centrality, betweenness centrality, closeness centrality, eigenvector centrality, PageRank, harmonic centrality, clustering coefficient, average neighbor degree, k-core, and load centrality. UMAP dimensionality reduction of centrality data (**Figure S10A**) classified nodes into three distinct categories. Network layout analysis (**Figure S10B**) revealed spatial distribution patterns: Category 2 nodes were located at the network's core; Category 1 nodes occupied intermediate layers; Category 3 **nodes** were predominantly peripheral.

Statistical analysis of terminology distribution (**Figure S10C**) showed that Category 2 nodes were primarily linked to therapeutic principles and methods, forming the core of the diagnostic-therapeutic network. Category 1 nodes were more associated with etiology and pathogenesis, emphasizing their theoretical relevance. In contrast, Category 3 nodes had the lowest proportions across all terminology types, reflecting their peripheral status.

Further comparisons of centrality metrics among categories (**Figure S10D**) confirmed that Category 2 nodes significantly outperformed the others, firmly establishing their core role. Category 1 nodes demonstrated intermediate importance, while Category 3 nodes consistently scored lowest. This stratification lays a critical foundation for extracting core features across categories and provides insights into the hierarchical structure and functional roles within the network.

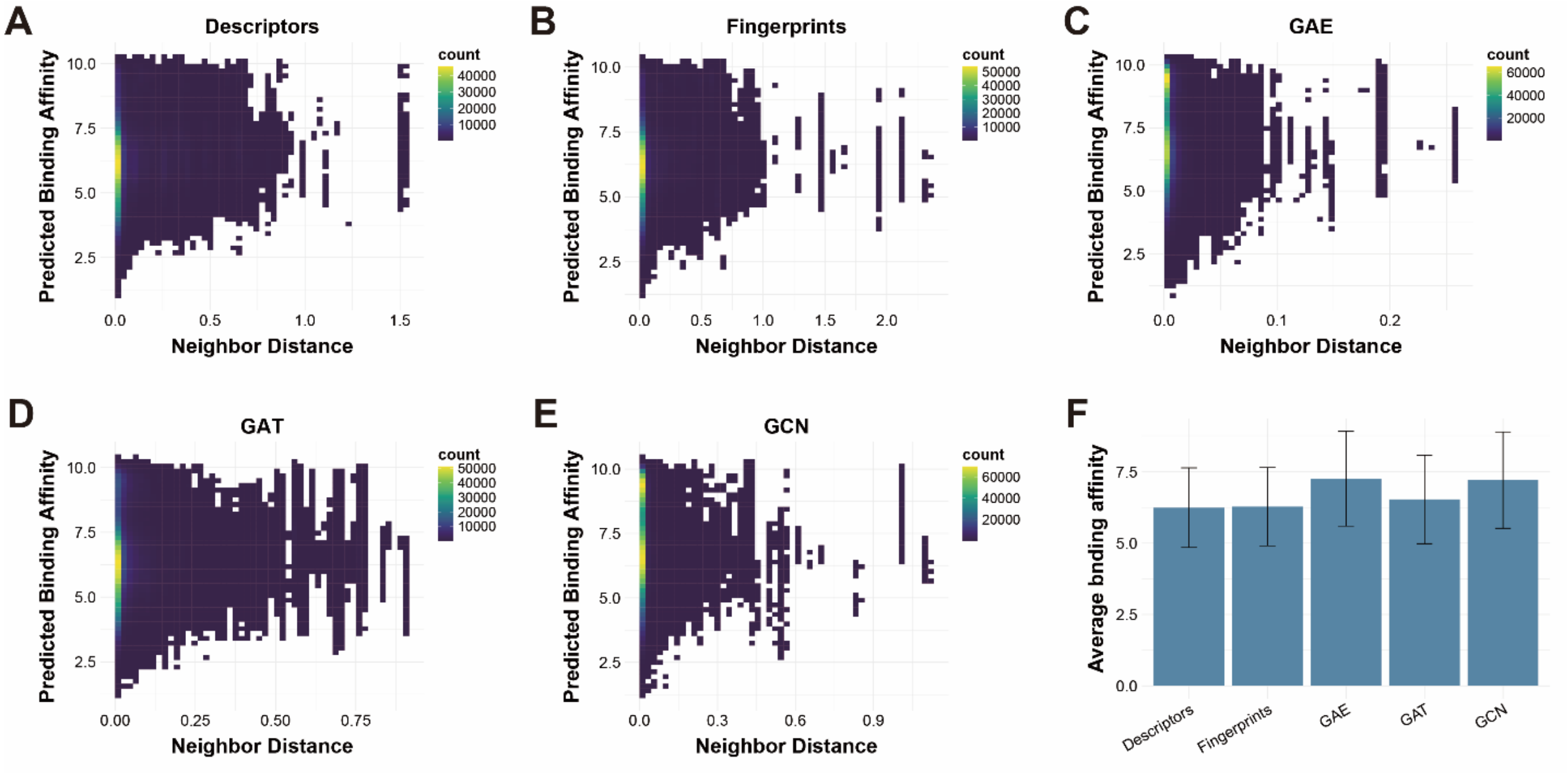

**Fig. S1. Relationship between neighbor distance and predicted binding affinity across different feature representation methods.** (A) Molecular descriptors (Descriptors). (B) Molecular fingerprints (Fingerprints). (C) Graph autoencoder (GAE). (D) Graph attention network (GAT). (E) Graph convolutional network (GCN). (F) Comparison of average binding affinity at short neighbor distances across different feature representation methods. Error bars represent standard deviations.

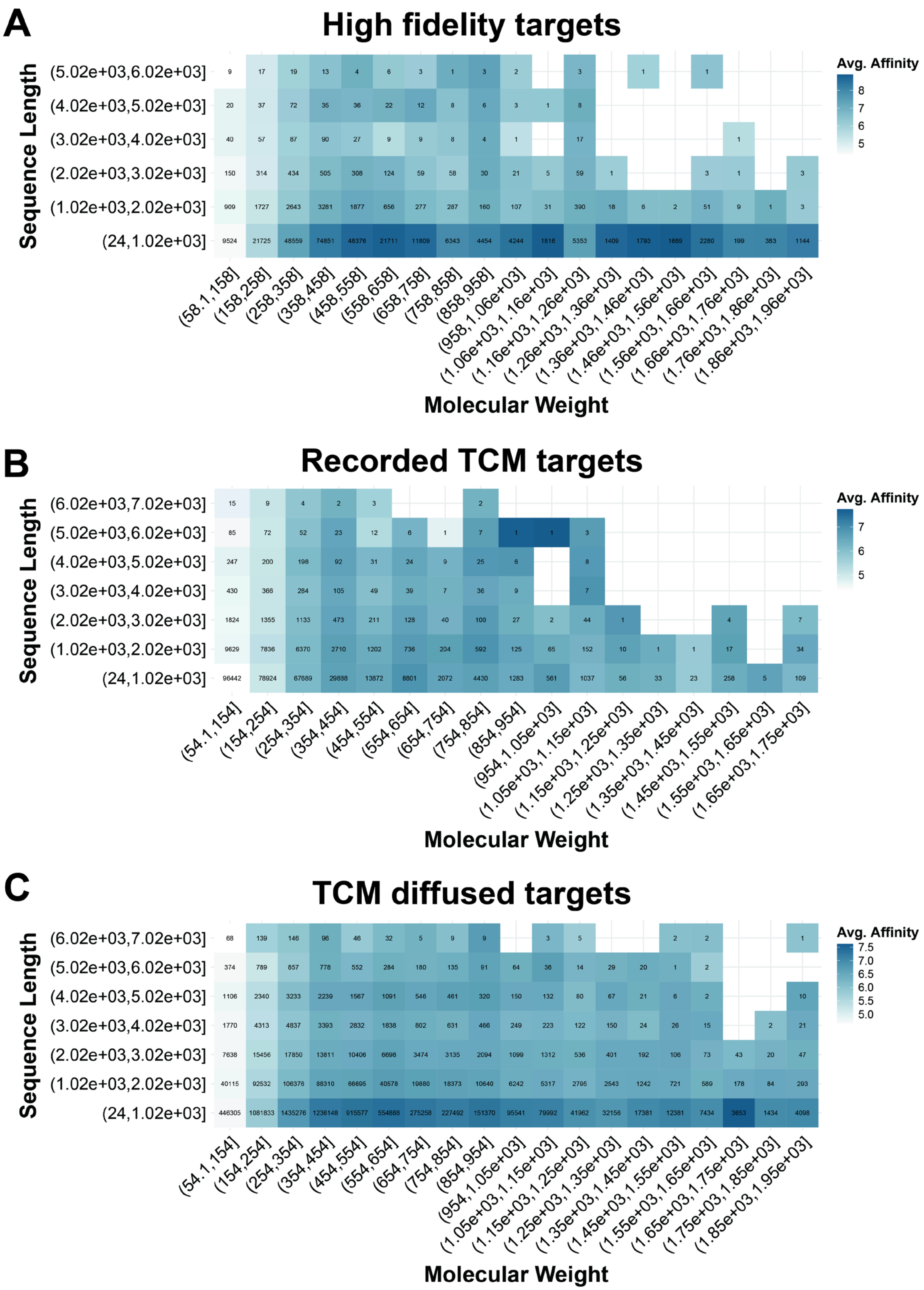

**Fig. S2. Heatmap of the effects of molecular weight and target sequence length on binding affinity.** Each grid shows the number of samples at that point. (A) High fidelity targets. (B) Recorded TCM targets. (C) TCM diffused targets.

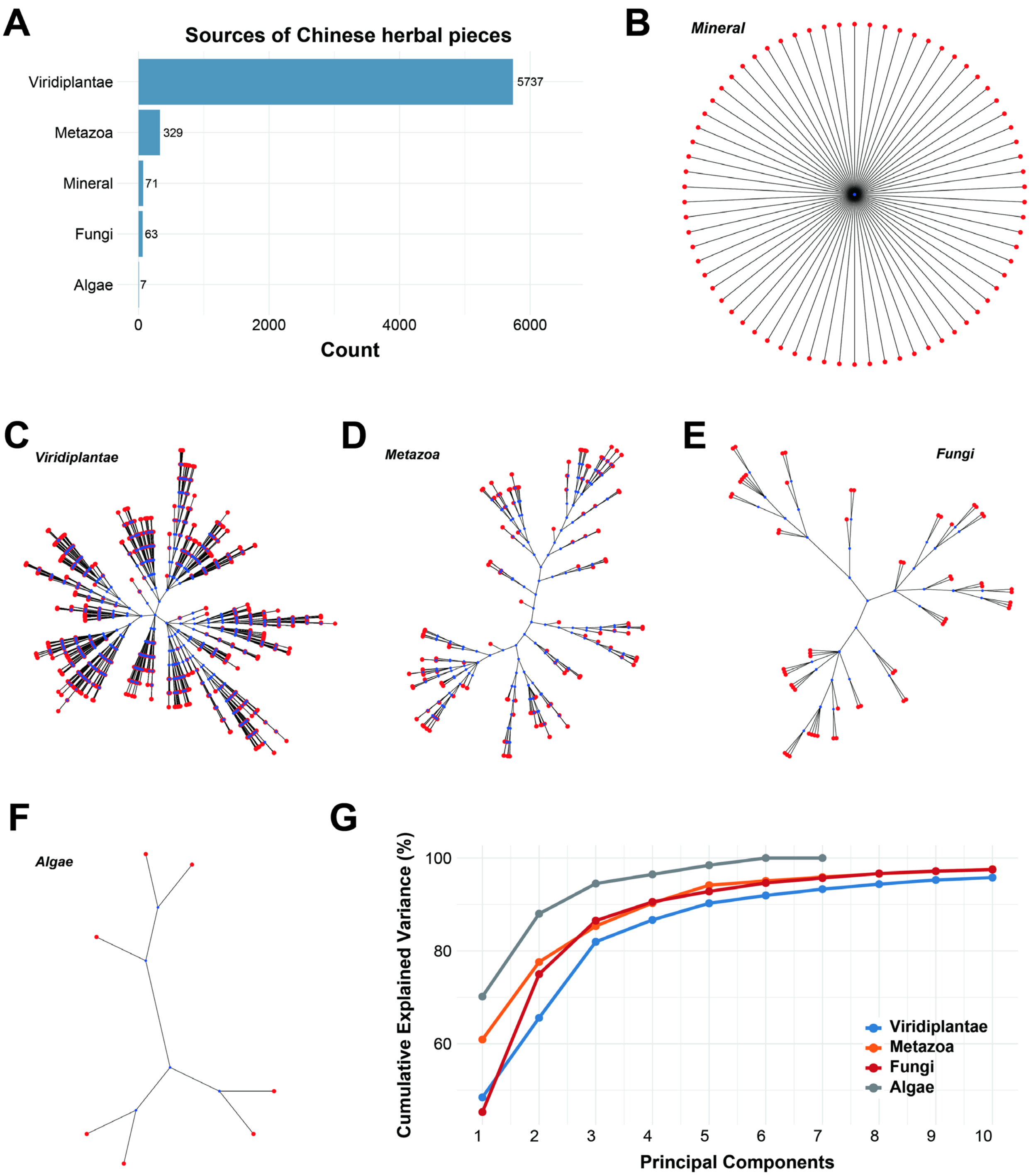

**Fig. S3. Classification and phylogenetic analysis of Chinese herbal pieces.** (A) Taxonomic distribution of Chinese herbal piece sources. (B–F) Phylogenetic trees of different source categories: (B) Mineral, (C) Viridiplantae, (D) Metazoa, (E) Fungi, and (F) Algae. (G) Principal component analysis (PCA) of source characteristics, showing cumulative explained variance.

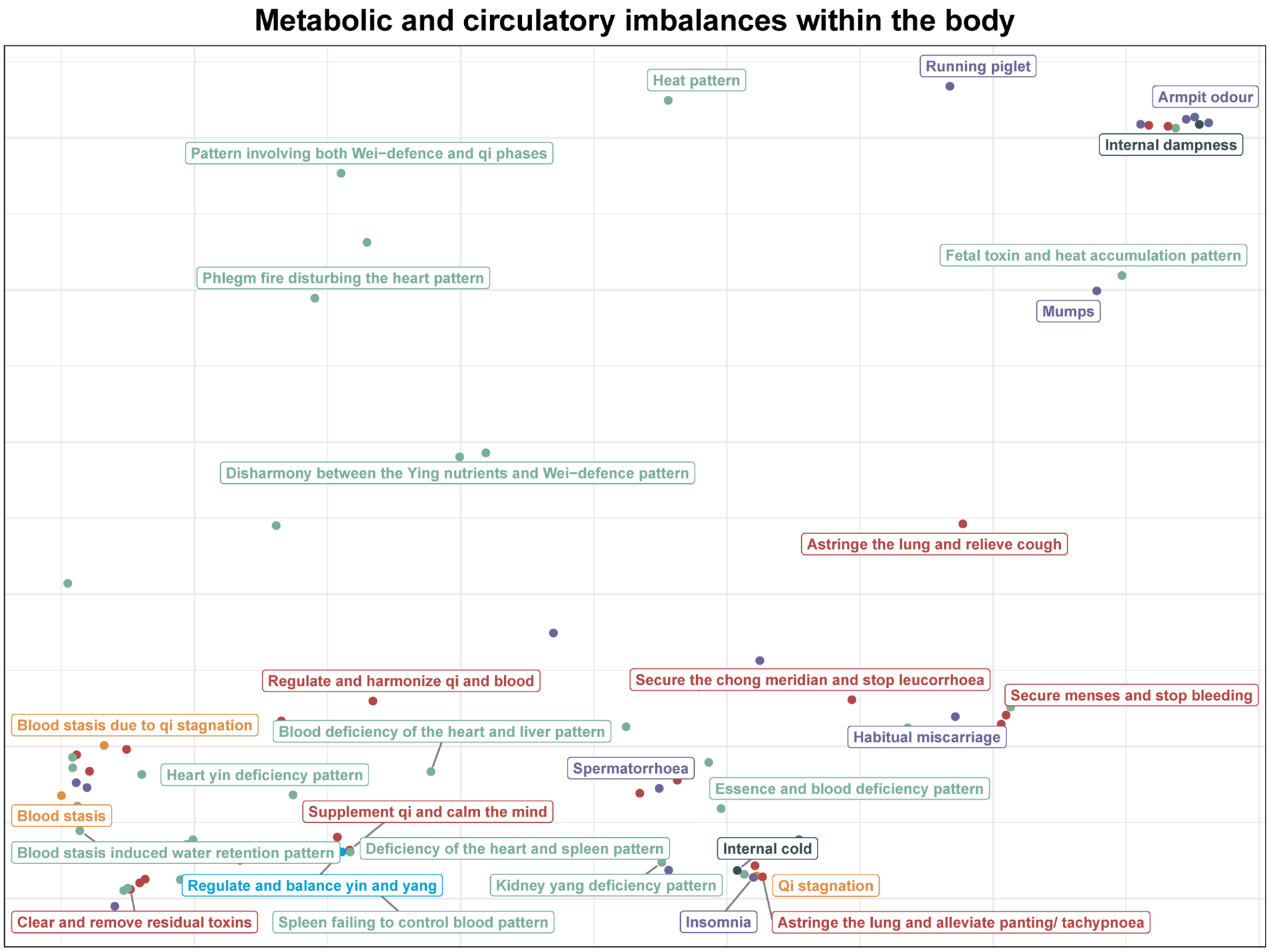

**Fig. S4. Metabolic and circulatory imbalances within the body.** This category represents conditions resulting from disruptions in metabolism and the circulatory system, characterized by pathologies such as qi stagnation, blood stasis, and internal dampness. Symptoms often include impaired digestion, fluid retention, and imbalances in yin and yang. TCM treatments emphasize warming yang, tonifying qi, and invigorating blood circulation to restore internal metabolic and circulatory harmony.

## Organ aging and immune deficiency

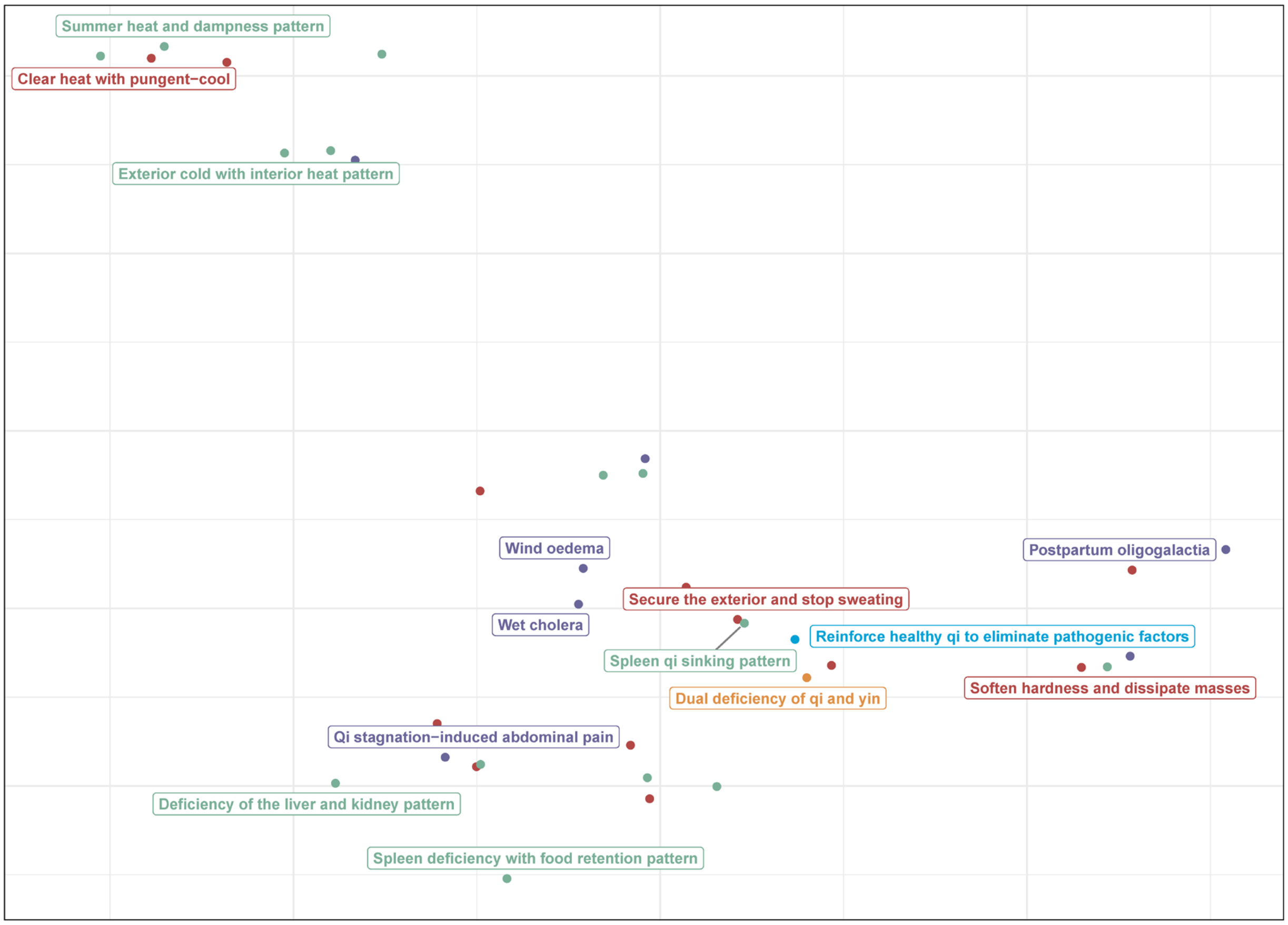

**Fig. S5. Organ aging and immune deficiency.** This category highlights conditions associated with aging and immune system decline, often presenting as qi deficiency, yin deficiency, and impaired organ function. Common syndromes include kidney yin deficiency, spleen qi sinking, and blood deficiency. TCM treatments focus on tonifying qi and yin, nourishing the kidneys, and enhancing overall immunity to slow organ degeneration and improve resilience.

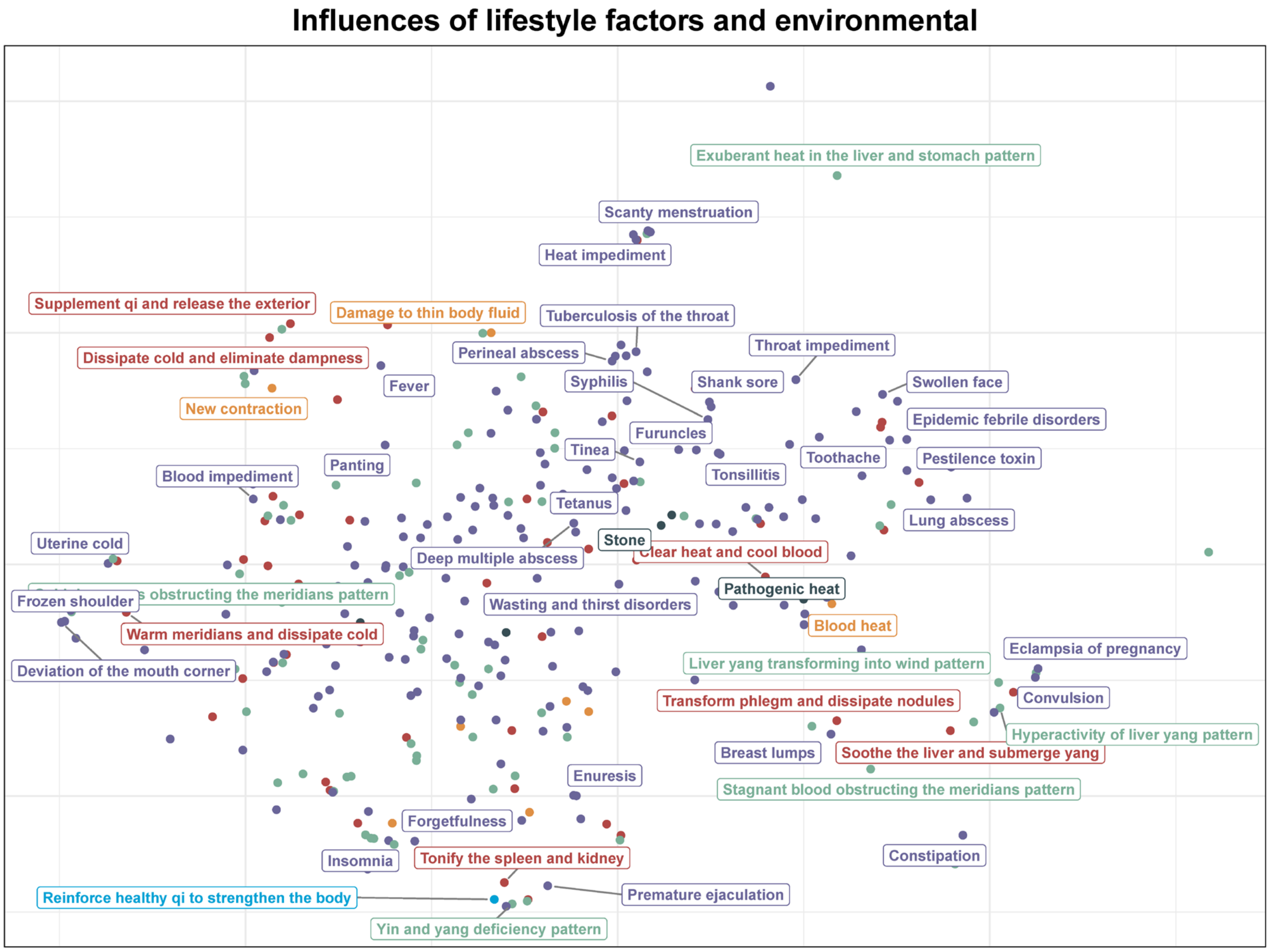

**Fig. S6. Influences of lifestyle and environmental conditions.** This category addresses the impact of lifestyle choices and environmental factors, including unhealthy diets and exposure to external cold or heat. Pathologies such as qi stagnation, blood heat, and dampness accumulation are common. TCM treatments target clearing heat, removing dampness, and invigorating blood flow to correct imbalances caused by external and lifestyle-related factors.

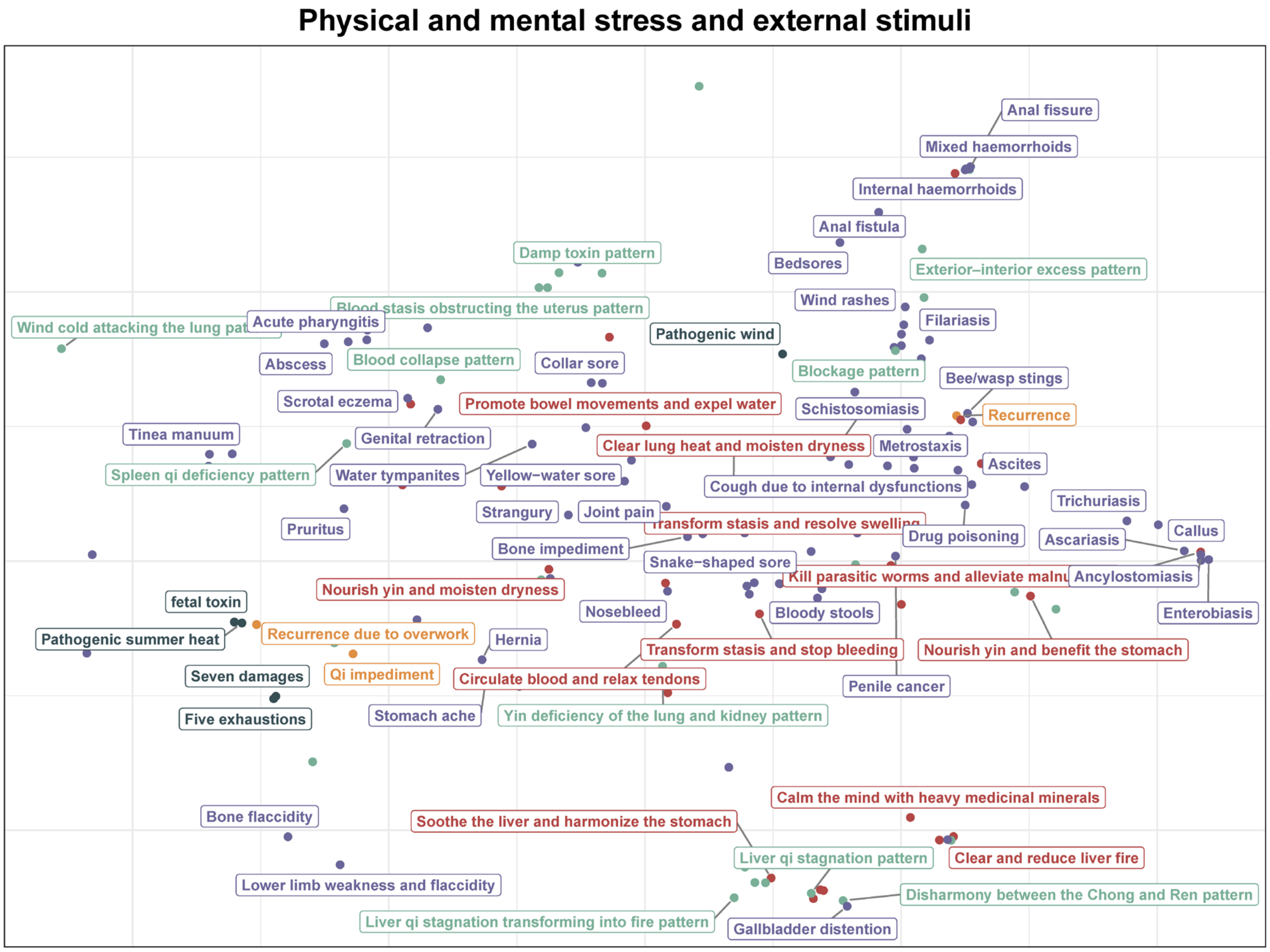

**Fig. S7. Physical and mental stress and external stimuli.** This category reflects conditions triggered by prolonged stress, overwork, or environmental stimuli, often manifesting as qi stagnation, heat, or wind invasion. Symptoms include insomnia, irritability, and fatigue. TCM treatments emphasize soothing the liver, relieving qi stagnation, and harmonizing the body to alleviate both physical and mental stressors.

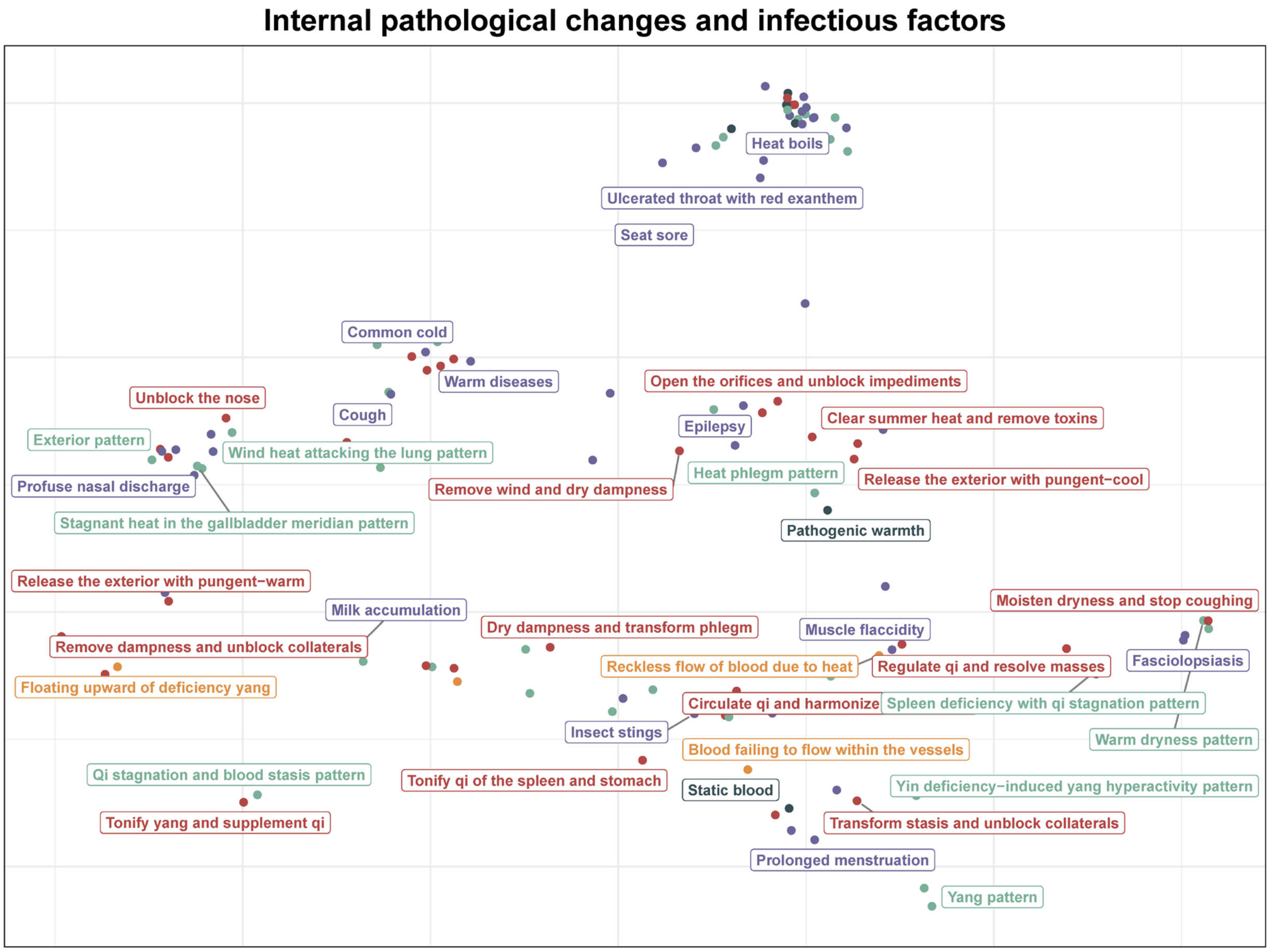

**Fig. S8. Internal pathological changes and infectious factors.** This category focuses on conditions arising from internal pathological changes and external infections. Key mechanisms include internal wind, blood stasis, and pathogenic warmth. Symptoms may involve heat boils, phlegm patterns, and chronic inflammation. TCM treatments integrate detoxification, clearing heat, and resolving phlegm to manage infections and restore internal balance.

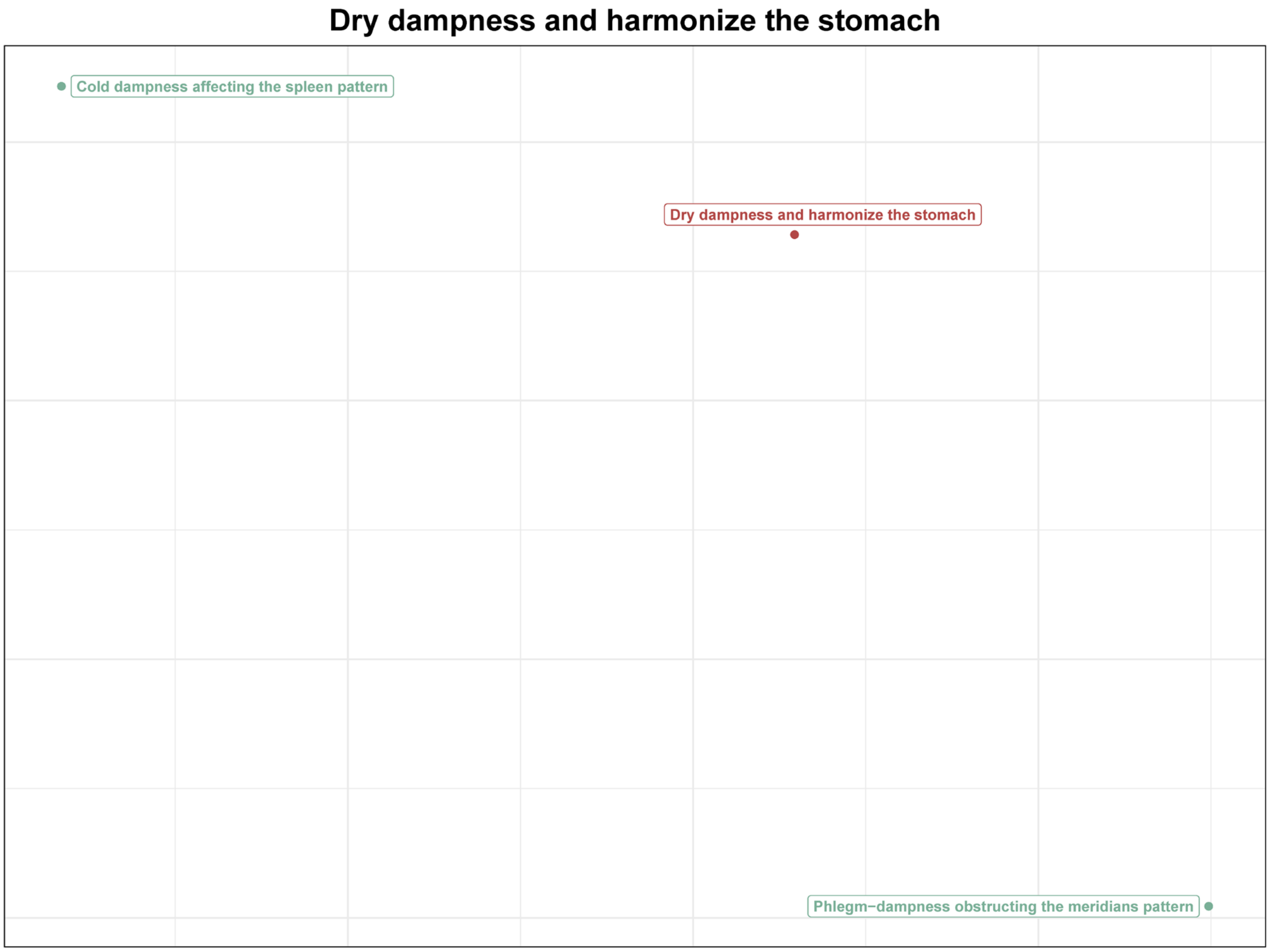

**Fig. S9. Dry dampness and harmonize the stomach.** This category, though less prominent, involves conditions linked to spleen and stomach dysfunction caused by dampness or phlegm retention. Symptoms include digestive issues and dampness obstruction. TCM treatments focus on drying dampness and harmonizing the stomach to restore digestive health and systemic balance.

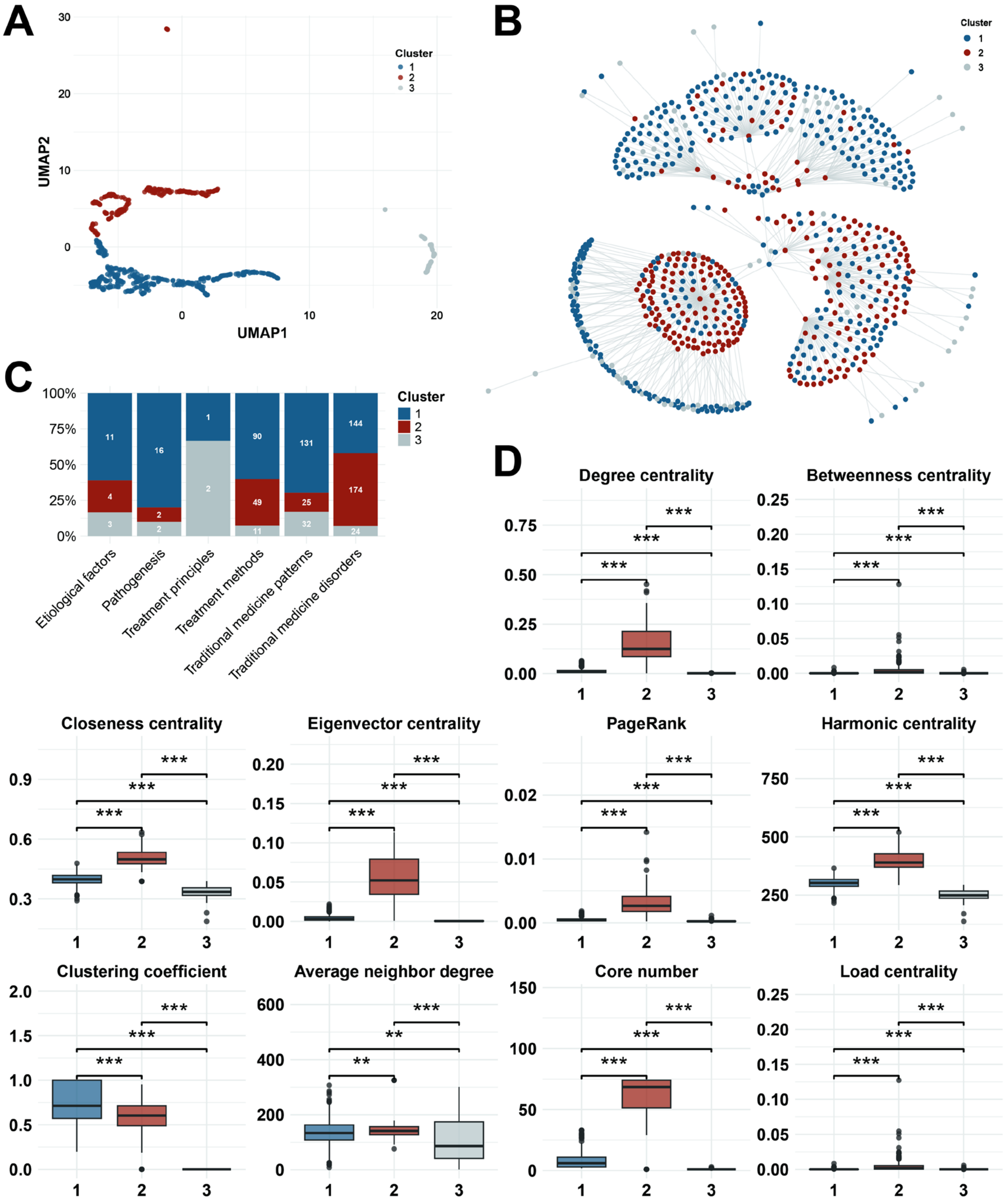

**Fig. S10. Core feature analysis of TCM diagnostic terminology.** (A) UMAP visualization of node importance based on centrality measures, including degree, betweenness, closeness, eigenvector, PageRank, harmonic, clustering coefficient, average neighbor degree, core number, and load centrality. (B) Network layout of nodes grouped by importance levels derived from centrality measures. (C) Proportional distribution of TCM diagnostic terminologies (etiological factors, pathogenesis, treatment methods, principles, and patterns) across node categories. (D) Centrality metric distributions for node categories, with Category 2 showing the highest centrality and core importance, followed by Category 1, and Category 3 being the least significant. $^{**}P < 0.01$; $^{***}P < 0.001$

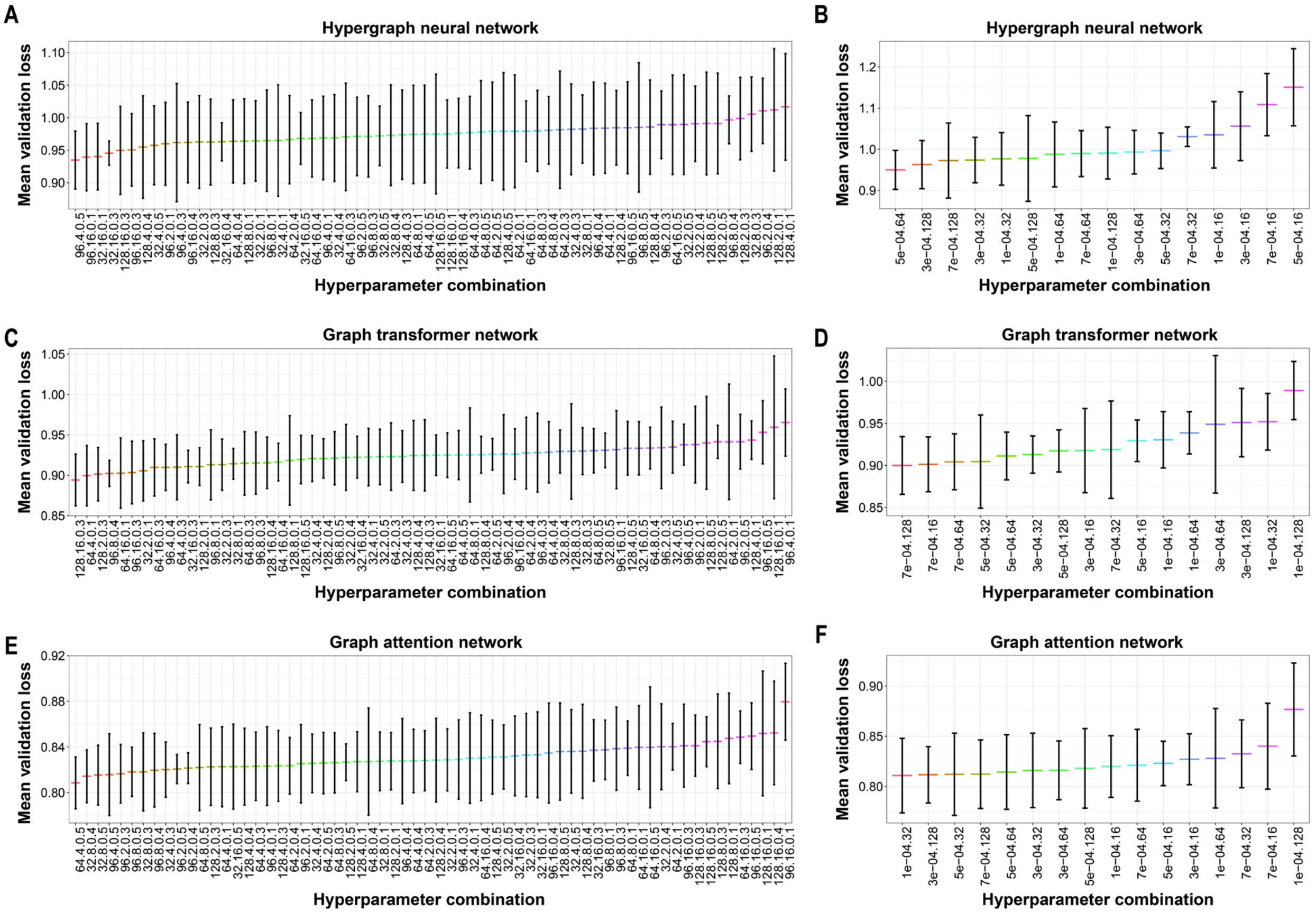

**Fig. S11. Hyperparameter optimization of three graph neural network models.** (A, B) Hypergraph neural network: Optimization of structural parameters (A) including hidden dimensions (hidden_dim), number of attention heads (num_heads), and dropout rate (dropout_rate), followed by training parameters (B) including learning rate (learning_rate) and batch size (batch_size). (C, D) Graph transformer network: Structural parameter optimization (C) and training parameter optimization (D), with mean validation loss showing the effectiveness of parameter combinations. (E, F) Graph attention network: Two-stage optimization of structural parameters (E) and training parameters (F), evaluated by mean validation loss. A two-stage grid search with 5-fold cross-validation was employed. Structural parameters were optimized first to establish the ideal architecture, followed by refinement of training parameters to stabilize training and enhance generalization, as indicated by reduced mean validation loss.

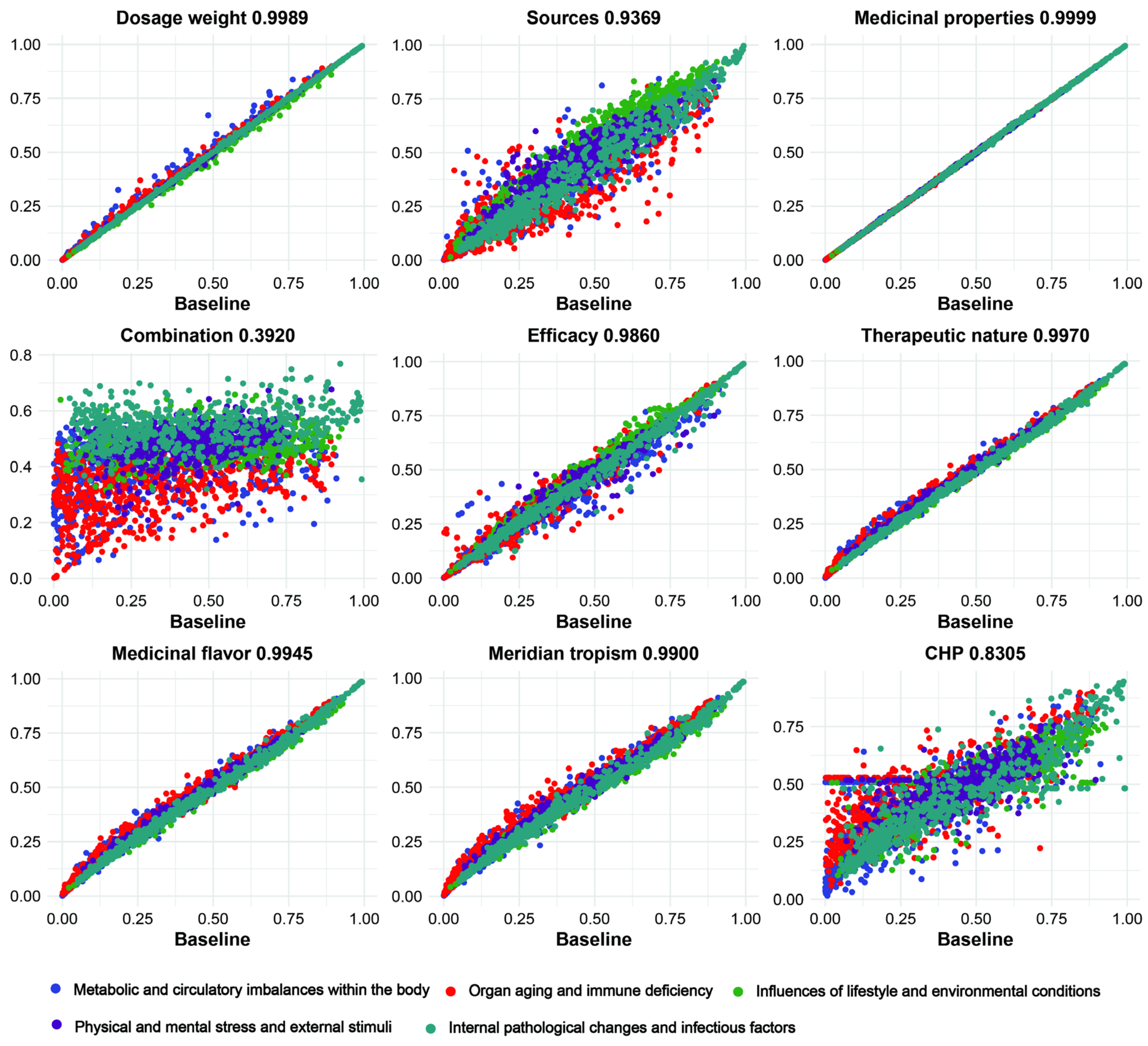

**Fig. S12. Analysis of prediction changes before and after feature and node zeroing.** Scatter plots illustrate the impact of feature and node zeroing on prediction results compared to the baseline using the graph attention network. Features: Dosage weight, sources, medicinal properties, combination, and efficacy. Nodes: Therapeutic nature, medicinal flavor, meridian tropism, and CHP (Chinese herbal pieces). Correlation coefficients between the zeroed and baseline predictions are indicated for each feature and node. Points are colored by TCM etiological types.

**Table S1. Performance comparison of graph neural network models across five major TCM etiological types**

| Model type | Metrics | Metabolic and circulatory imbalances within the body | | | Organ aging and immune deficiency | | | Influences of lifestyle and environmental conditions | | | Physical and mental stress and external stimuli | | | Internal pathological changes and infectious factors | | |
|---|---|---|---|---|---|---|---|---|---|---|---|---|---|---|---|---|
| | | Cal | Val | Test | Cal | Val | Test | Cal | Val | Test | Cal | Val | Test | Cal | Val | Test |
| Graph transformer network | Precision | 0.166 | 0.1751 | 0.1111 | 0.1313 | 0.1347 | 0.1111 | 0.7082 | 0.6961 | 0.7258 | 0.1606 | 0.1538 | 0.1483 | 0.5665 | 0.5558 | 0.5774 |
| | Recall | 0.8037 | 0.8214 | 0.7692 | 0.8352 | 0.8193 | 0.675 | 0.772 | 0.7819 | 0.7989 | 0.507 | 0.5179 | 0.5167 | 0.7924 | 0.7466 | 0.7806 |
| | F1 Score | 0.2752 | 0.2887 | 0.1942 | 0.2269 | 0.2313 | 0.1908 | 0.7387 | 0.7365 | 0.7606 | 0.244 | 0.2372 | 0.2305 | 0.6607 | 0.6372 | 0.6638 |
| | AUC | 0.8259 | 0.8445 | 0.8168 | 0.7962 | 0.7852 | 0.7446 | 0.733 | 0.6971 | 0.7641 | 0.6672 | 0.6713 | 0.6178 | 0.8336 | 0.8187 | 0.8241 |
| | Accuracy | 0.7314 | 0.7218 | 0.7243 | 0.6349 | 0.6301 | 0.6196 | 0.687 | 0.6768 | 0.7093 | 0.6826 | 0.6948 | 0.6561 | 0.7521 | 0.7447 | 0.7425 |
| Hypergraph neural network | Precision | 0.3133 | 0.2768 | 0.1947 | 0.2621 | 0.2218 | 0.1655 | 0.7895 | 0.7574 | 0.7578 | 0.2559 | 0.1843 | 0.1961 | 0.6758 | 0.6241 | 0.6301 |
| | Recall | 0.9074 | 0.7381 | 0.8462 | 0.8938 | 0.7349 | 0.575 | 0.7368 | 0.721 | 0.7011 | 0.8279 | 0.6518 | 0.6667 | 0.7978 | 0.7057 | 0.7041 |
| | F1 Score | 0.4658 | 0.4026 | 0.3165 | 0.4053 | 0.3408 | 0.257 | 0.7622 | 0.7388 | 0.7284 | 0.391 | 0.2874 | 0.303 | 0.7318 | 0.6624 | 0.6651 |
| | AUC | 0.9558 | 0.8869 | 0.8994 | 0.9386 | 0.8028 | 0.7766 | 0.8209 | 0.7674 | 0.7841 | 0.8627 | 0.7323 | 0.7317 | 0.8909 | 0.8295 | 0.8445 |
| | Accuracy | 0.868 | 0.8494 | 0.8422 | 0.8318 | 0.8069 | 0.7791 | 0.7366 | 0.7054 | 0.6977 | 0.7394 | 0.7038 | 0.6944 | 0.8219 | 0.784 | 0.7691 |
| | Specificity | 0.8653 | 0.8576 | 0.842 | 0.8275 | 0.8121 | 0.7936 | 0.7364 | 0.6841 | 0.6929 | 0.7295 | 0.709 | 0.6974 | 0.8324 | 0.8175 | 0.8005 |
| Graph attention network | Precision | 0.2241 | 0.2207 | 0.1489 | 0.2191 | 0.215 | 0.1825 | 0.7465 | 0.7289 | 0.7647 | 0.219 | 0.1802 | 0.2113 | 0.6182 | 0.6104 | 0.6203 |
| | Recall | 0.8407 | 0.7857 | 0.8077 | 0.7656 | 0.759 | 0.625 | 0.7618 | 0.7691 | 0.7471 | 0.714 | 0.6518 | 0.75 | 0.7809 | 0.7384 | 0.75 |
| | F1 Score | 0.3539 | 0.3446 | 0.2515 | 0.3407 | 0.3351 | 0.2825 | 0.7541 | 0.7484 | 0.7558 | 0.3352 | 0.2824 | 0.3297 | 0.6901 | 0.6683 | 0.679 |
| | AUC | 0.9004 | 0.8842 | 0.8918 | 0.8736 | 0.8347 | 0.8106 | 0.7939 | 0.765 | 0.8013 | 0.7757 | 0.7341 | 0.7589 | 0.8644 | 0.8393 | 0.8366 |
| | Accuracy | 0.8052 | 0.7946 | 0.7924 | 0.8099 | 0.7954 | 0.789 | 0.7152 | 0.7013 | 0.7209 | 0.7138 | 0.6964 | 0.696 | 0.7864 | 0.7799 | 0.7691 |
| | Specificity | 0.8028 | 0.7953 | 0.7917 | 0.813 | 0.7981 | 0.8007 | 0.6527 | 0.6085 | 0.685 | 0.7138 | 0.7009 | 0.69 | 0.7889 | 0.7977 | 0.7783 |

The dataset comprises 6,080 samples, divided into 4,256 training graphs (Cal), 1,222 validation graphs (Val), and 602 test graphs (Test), following a 7:2:1 split. Positive sample proportions for the five major TCM etiological types are as follows:

Metabolic and circulatory imbalances within the body: Training (6.34%), Validation (6.87%), Test (4.32%);

Organ aging and immune deficiency: Training (6.41%), Validation (6.79%), Test (6.64%);

Influences of lifestyle and environmental conditions: Training (57.31%), Validation (57.77%), Test (57.81%);

Physical and mental stress and external stimuli: Training (10.10%), Validation (9.17%), Test (9.97%);

Internal pathological changes and infectious factors: Training (30.45%), Validation (30.03%), Test (32.56%).

**Supplemental Data**

**Data S1.** Overview of TCM-MKG data sources and key information tables.

**Data S2.** Hyperparameter optimization and evaluation of graph neural network models for Chinese herbal formulas classification.

**Data S3.** Feature and node contribution analysis using zeroing method.

**Data S4.** Chinese herbal formulas for COVID-19 management: collection and composition.

**Data S5.** Model prediction results and attention weights for Chinese herbal formulas in COVID-19 management.

**Data S6.** Analysis of frequency and attention weights of Chinese herbal pieces and herb pairs in COVID-19 formulas.

**Data S7.** Component-target network for Radix Astragali and associated KEGG pathways